\documentclass[sigconf,nonacm]{acmart}
\usepackage{amsmath}
\usepackage{amsfonts}
\usepackage{graphicx}
\usepackage{multirow}
\usepackage{booktabs}
\usepackage{amsmath}
\usepackage{amsfonts}
\usepackage{xcolor}
\usepackage{url}
\usepackage{subcaption}
\usepackage{rotating}
\usepackage{wrapfig}
\usepackage{array}
\usepackage{xspace}
\usepackage{hyperref}
\usepackage{comment}

\newtheorem{defn}{Definition}
\newcommand{\ssnp}{\textit{SSNP}\xspace}
\DeclareMathOperator{\poolssnp}{pool_{SSNP}}
\DeclareMathOperator{\poolsnp}{pool_{SNP}}
\newcommand{\poolssnpT}{$\poolssnp$\xspace}
\newcommand{\first}{\textcolor[rgb]{0.9, 0.17, 0.31}}
\newcommand{\second}{\textcolor[rgb]{0.0, 0.0, 1.0}}
\newcommand{\third}{\textcolor[rgb]{0.55, 0.0, 0.55}}
\DeclareMathOperator{\pool}{pool}

\AtBeginDocument{%
  \providecommand\BibTeX{{%
    \normalfont B\kern-0.5em{\scshape i\kern-0.25em b}\kern-0.8em\TeX}}}
    
\begin{document}
\title{Stochastic Subgraph Neighborhood Pooling for Subgraph Classification}
%






\author{Shweta Ann Jacob}
\email{shweta.jacob@ontariotechu.net}
\affiliation{%
  \institution{Ontario Tech University}
  \city{Oshawa}
  \state{Ontario}
  \country{Canada}
}

\author{Paul Louis}
\email{paul.louis@ontariotechu.net}
\affiliation{%
  \institution{Ontario Tech University}
  \city{Oshawa}
  \state{Ontario}
  \country{Canada}
}

\author{Amirali Salehi-Abari}
\email{abari@ontariotechu.ca}
\affiliation{%
  \institution{Ontario Tech University}
  \city{Oshawa}
  \state{Ontario}
  \country{Canada}
}
%
%
%

\renewcommand{\shortauthors}{S. Jacob et al.}

\begin{abstract}
Subgraph classification is an emerging field in graph representation learning where the task is to classify a group of nodes (i.e., a subgraph) within a graph. Subgraph classification has applications such as predicting the cellular function of a group of proteins or identifying rare diseases given a collection of phenotypes. Graph neural networks (GNNs) are the de facto solution for node, link, and graph-level tasks but fail to perform well on subgraph classification tasks. Even GNNs tailored for graph classification are not directly transferable to subgraph classification as they ignore the external topology of the subgraph, thus failing to capture how the subgraph is located within the larger graph. The current state-of-the-art models for subgraph classification address this shortcoming through either labeling tricks or multiple message-passing channels, both of which impose a computation burden and are not scalable to large graphs. To address the scalability issue while maintaining generalization, we propose \textit{Stochastic Subgraph Neighborhood Pooling (\ssnp)}, which jointly aggregates the subgraph and its neighborhood (i.e., external topology) information without any computationally expensive operations such as labeling tricks. To improve scalability and generalization further, we also propose a simple data augmentation pre-processing step for \ssnp that creates multiple sparse views of the subgraph neighborhood. We show that our model is more expressive than GNNs without labeling tricks. Our extensive experiments demonstrate that our models outperform current state-of-the-art methods (with a margin of up to 2\%) while being up to 3$\times$ faster in training.
\end{abstract}


\settopmatter{printfolios=true}
\maketitle    
\section{Introduction}
Graph-structured data is prevalent in many domains such as social networks, biological networks (e.g., protein-interaction networks), or technological networks (e.g., information networks or computer networks). Structural properties of graph data have been exploited for drug repurposing/discovery \cite{morselli2021network,jiang2021could}, recommender systems \cite{wang2014friendbook,ying2018graph,salehi2015preference}, medical diagnosis \cite{alsentzer2022deep}, peer assessment \cite{namanloo2022improving}, and many more. Graph representation learning has continuously progressed in recent years with the advent of more expressive graph neural networks (GNNs) \cite{kipf2017semi,hamilton2017inductive,velivckovic2017graph,xu2018powerful}, 
focusing on various downstream tasks such as node classification \cite{rossi2020sign}, link prediction \cite{zhang2018link}, and graph classification \cite{zhang2018end}.

Subgraph classification is an emerging problem in graph representation learning where one intends to  predict the properties associated with a group of nodes (i.e., a \textit{subgraph}) of the larger observed \textit{base} graph \cite{alsentzer2020subgraph,wang2021glass}. Subgraph classification finds application in various domains such as finding toxic (or violence-inciting) communities in social networks, drug discovery, group recommendation, diagnosis of rare diseases, and many others. As subgraphs may contain any number of nodes ranging from one node to all nodes of the base graph, typical downstream tasks (e.g., node classification, link prediction, or graph classification) can be considered as specific instances of subgraph classification. 

Subgraph classification, as a more general problem, requires solutions that can learn, combine, and contrast topological properties and the connectivity between the nodes within and outside the subgraph. Learning these complex intra-connectivity and inter-connectivity patterns of the subgraph and the base graph renders this problem challenging. As a result, existing GNN models that perform well on node classification, link prediction, and graph classification does not perform well on subgraph classification \cite{wang2021glass}. Also, learning solely on segregated subgraphs that ignore the topology of the base graph is shown to be ineffective \cite{wang2021glass}, thus underpinning the importance of the global topology of the base graphs for the subgraph classification task. Recent state-of-the-art work (e.g., GLASS \cite{wang2021glass} and SubGNN \cite{alsentzer2020subgraph}) alleviates this shortcoming of the lack of global topology information through the use of labeling tricks \cite{wang2021glass} or artificially-crafted message passing channels \cite{alsentzer2020subgraph}.

While GLASS \cite{wang2021glass} and SubGNN \cite{alsentzer2020subgraph} enhance the expressiveness of subgraph embeddings, their deployed approaches of labeling tricks and additional artificially-crafted message passing channels are computationally intensive, especially when dealing with larger (sub)graphs. In some cases, these computational bottlenecks have made these approaches require some careful hyperparameter tuning. For example, the performance of the max-zero-one labeling trick in GLASS is sensitive to the batch size and therefore, requires extensive and careful hyperparameter tuning of the batch size. To overcome the computational overhead of GLASS and SubGNN, it is essential to devise a model that can learn the interactions between subgraph nodes and the external nodes without any computationally-costly subgraph-level operations.

We introduce a simple computational-friendly model for subgraph classification that does not use any labeling trick or artificially fabricated computationally expensive message-passing channels. Operating on the original graph, our model does not require any subgraph extractions. We first utilize transformation layers on the node features of all nodes in the base graph for 
dimensionality reduction and node feature smoothing/refinement. The transformation layers can be message-passing layers such as GCN \cite{kipf2017semi}, GraphSAGE \cite{hamilton2017inductive}, GIN \cite{xu2018powerful}, or a simple graph structure-agnostic model such as MLP. Then, for each subgraph, we aggregate the node features of the subgraph and its neighborhood through our proposed \textit{Stochastic Subgraph Neighborhood Pooling (\ssnp)} to generate the subgraph embedding, and consequently the subgraph classification output. The addition of subgraph neighborhood information in our pooling function enhances the expressiveness of subgraph embeddings by capturing their external topology within a base graph. We show that our model is more expressive than a \textit{plain GNN} (i.e., a simple graph neural network such as GCN \cite{kipf2017semi} without any labeling tricks). To prevent neighborhood explosion for large graphs and keep computation under control, our \ssnp uses random walks to sample the neighborhood of each subgraph. As a data augmentation strategy, our neighborhood sampling method can be conducted multiple times in a pre-processing stage to create multiple sparse views of the subgraph neighborhood.  We conduct comprehensive experiments on real-world datasets to show the performance and scalability of our model against various baselines including the current state-of-the-art GLASS \cite{wang2021glass}. In all datasets (except one), our model outperforms others with a gain of up to 2\% while having a speedup of up to 3$\times$ compared to GLASS. Experimental results on real-world datasets demonstrate our model is effective, yet simple and computationally efficient. Moreover, the utilization of subgraph neighborhoods in the pooling layer enhances the power of the subgraph representations without the requirement for any labeling trick.

\section{Related Work}
We discuss related work relevant to our line of research. We first review recent work on GNNs that have been successful in other downstream tasks, and then discuss prominent work in subgraph classification and scalability.

\vskip 1mm
\noindent \textbf{Graph Neural Networks.} 
The early work in graph representation learning was \textit{shallow encoders} such as DeepWalk \cite{perozzi2014deepwalk} and node2vec \cite{grover2016node2vec}. DeepWalk uses random walks to understand the neighborhood around each node and encodes the sequence of random walks as node representations. An extension of Deepwalk, node2vec \cite{grover2016node2vec} introduced two hyperparameters to control the trade-off between breadth-first and depth-first exploration for random walks. Due to the inapplicability of such methods in inductive settings and their negligence of nodal features, message-passing graph neural networks (MPGNNs) \cite{bruna2013spectral,defferrard2016convolutional} were introduced and popularized. GCN \cite{kipf2017semi}, one of the most popular MPGNN models, iteratively updates nodal representations by aggregating messages from its neighbors. However, GCN suffers from the exploding neighborhood problem with a high number of node feature updates. To overcome this, GraphSAGE \cite{hamilton2017inductive} used a neighborhood sampling method during message passing that allowed for the model to be inductive as well as scalable. The continuing body of research on MPGNNs has two main directions: (i) improving the message passing scheme by computing different weights or attention \cite{velivckovic2017graph} for node feature aggregation in a neighborhood; or (ii) enhancing the expressiveness of graph neural networks by applying a multi-layer perceptron to the nodes after message passing \cite{xu2018powerful}. 

\vskip 1mm
\noindent \textbf{Subgraph Representation Learning.} 
Despite their success on the node and graph classification tasks, MPGNNs fail to uniquely capture pairwise nodal interactions \cite{wang2023neural,chamberlain2023graph}. To circumvent this, SEAL \cite{zhang2018link} converts link prediction to a graph classification problem by extracting enclosing subgraphs around each pair of target nodes, which are then used to predict the existence/absence of the link. To understand the structure of the enclosing subgraph, SEAL injects distance information as nodal features of the subgraph nodes using double-radius node labeling (based on distances of nodes in a subgraph to the target nodes). Labeling tricks (e.g., double-radius) are shown to allow the underlying model to learn the dependency between the target nodes in their neighborhood subgraphs \cite{zhang2021labeling}. Subgraph representation learning approaches (SGRLs), by using the enclosing subgraphs around the target pair and labeling tricks,  have enhanced the expressive power of MPGNNs for link prediction \cite{lidistance,zhang2018link,cai2020multi,pan2022neural}. This success even has extended to other downstream tasks. For example, shaDow-GNN \cite{zeng2021decoupling} extract $K$-hop subgraphs around each node and operate on them for node classification. Similarly, NGNN \cite{zhang2021nested} aggregates the node features in the $K$-hop rooted subgraphs around each node to increase the expressiveness of representations for graph classification. Recently, $\text{I}^2$-GNNs \cite{huang2023boosting} extended NGNN by using both labeling tricks and subgraph-level information to improve graph classification and cycle counting in graphs. $\text{I}^2$-GNNs does this by labeling the root node and one of its neighbors in the $K$-hop subgraph before message passing, thereby increasing the nodal representational power of MPGNNs.

\begin{figure*}[t]
  \centering
  \includegraphics[width=\textwidth]{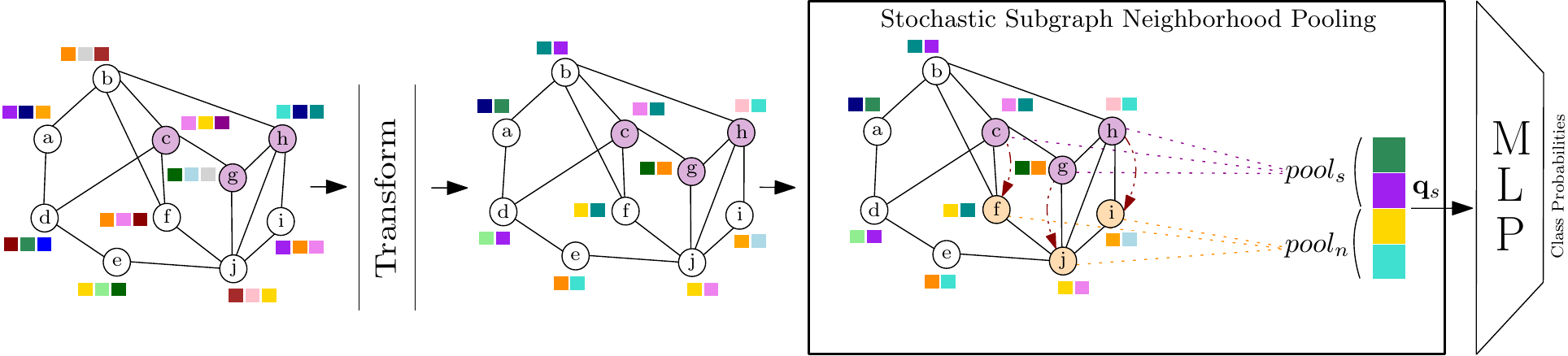}
  \caption{Architecture of our model. Subgraph nodes are shaded in purple. The initial node features are transformed using transformation layers such as Nested Network convolutions, GCN convolutions, or MLP. The stochastic subgraph neighborhood pooling \poolssnpT is applied in multiple steps. The subgraph neighborhood nodes (shaded in brown) are sampled by rooted random walks (red dashed arrows). The subgraph and its sampled neighborhood are separately pooled by $pool_s$ and $pool_n$, which are simple graph pooling operators (e.g., mean, sum, etc.). The pooling outputs are concatenated to form the subgraph representation $\mathbf{q}_s$, which is passed to an MLP for generating class probabilities.}
  \label{fig:architecture}
  \end{figure*}
  
\vskip 1mm
\noindent \textbf{Subgraph Classification.}
Subgraph classification \cite{alsentzer2020subgraph,wang2021glass} is an emerging problem, which extends subgraph representation learning. SubGNN \cite{alsentzer2020subgraph} samples anchor patches from the base graph and propagates messages between anchors to the subgraph in multiple channels to learn the internal and external topologies of subgraphs. 
The anchor patches are sampled to encode properties such as neighborhood, position, and structure of a subgraph in the base graph. 
While SubGNN learns the different topological properties of the subgraphs, sampling of channel-specific anchor patches followed by propagation is computationally expensive. The state-of-the-art GLASS \cite{wang2021glass} uses the zero-one labeling trick \cite{zhang2021labeling} to differentiate between the internal and external nodes of a subgraph and thereby encode various topological properties of the subgraph. GLASS further modifies zero-one labeling to max zero-one labeling to enable mini-batch training.  
Sub2Vec \cite{adhikari2018sub2vec}, as a subgraph embedding model, is deployed for community detection and graph classification; however, it can be adapted for the subgraph classification task. Sub2Vec samples random walks from a node within each subgraph to learn its structure and neighborhood. This information is then fed into Paragraph2Vec \cite{le2014distributed} to create the final subgraph embeddings. Some recent work on subgraph classification includes PADEL \cite{liu2023position} and Subgraph-To-Node (S2N) \cite{kim2022efficient}. PADEL uses data augmentation and contrastive learning techniques along with position encodings of nodes during message passing. Subgraph-To-Node (S2N) \cite{kim2022efficient} translates the subgraphs to nodes to coarsen the base graph, and casts the subgraph classification task to a node classification task. 

\vskip 1mm
\noindent \textbf{Scalability of SGRLs.}
SGRLs are computationally demanding due to the extraction of subgraphs around target nodes, applying labeling tricks, and running GNNs on each subgraph. To address this computational bottleneck, a recent line of research has emerged in scalable SGRLs. SaGNN \cite{wang2022towards} enhances the expressiveness of a graph neural network by aggregating node representations in the rooted subgraph around each node in the base graph to make the model subgraph-aware. SaGNN does not use the subgraphs for message passing but only at the aggregation step. 
ScaLed \cite{louis2022sampling} extends SEAL by sparsifying the enclosing subgraphs around the target nodes to reduce the computational overhead associated with large subgraph sizes. The random-walk-induced subgraphs approximate the enclosing subgraphs without substantial performance compromises. SUREL \cite{yin2022algorithm}, similar to ScaLed, uses pre-computed random walks around each pair of nodes to approximate the subgraph, however, does not use MPGNNs. ELPH/BUDDY \cite{chamberlain2023graph} uses computationally-light algorithms to derive subgraph sketches for approximating the neighborhood overlap and unions around target nodes for faster message-passing without explicit subgraph extractions.
S3GRL \cite{louis2023simplifying} models speed up the training and inference of SGRL methods by simplifying the underlying GNN message-passing and aggregation steps. S3GRL does this by removing the non-linearity in-between graph convolutions, thus allowing precomputation of the subgraph-level message passing, and consequently faster training and inference.

\section{Preliminaries}
Let $G=(V, E)$ represent a simple, undirected graph where $V=\{1, \dots,n\}$ is the set of nodes (e.g., users, scientists, articles, proteins, etc.), and $E \subseteq V\times V$ represents the edge set (e.g., friendships, collaborations, citations, interactions, etc.). We sometimes represent $G$ by the adjacency matrix $\mathbf{A} \in \mathbb{R}^{n\times n}$ where $a_{ij}=1$ if an edge exists between nodes $i$ and $j$, and $0$ otherwise. We also assume each node $i \in V$ possesses a $d$-dimensional feature $\mathbf{x}_i \in \mathbb{R}^d$ (e.g., user information, research profile, keywords, protein characteristics). We sometimes stack all nodal features, row-by-row in the feature matrix $\mathbf{X}$ whose $i$-th row contains $\mathbf{x}_i$.  We consider a subgraph $S=(V_S, E_S)$ in base graph $G$ where $V_S \subseteq V$ and $E_S \subseteq (V_S \times V_S) \cap E$. 

\vskip 1.5mm
\noindent \textbf{Subgraph Classification Problem}. The goal is to learn a mapping function $f(G,\mathbf{X}, S)$ which takes the base graph $G$, its node feature matrix $\mathbf{X}$, and a subgraph $S$ as an input, and outputs the subgraph class label $y \in \{1,\dots, C\}$, where $C$ is the number of classes. The class labels of the subgraph could represent the toxic friendship communities, cellular functions (e.g., metabolism, development, etc), or metabolic/neurological disorders.


\section{Stochastic Subgraph Neighborhood Pooling (\ssnp)}
We first discuss the various components of our proposed solution for subgraph classification. We then detail an important part of this solution, our proposed Stochastic Subgraph Neighborhood Pooling (\ssnp).


Our proposed solution for subgraph classification is depicted in Figure \ref{fig:architecture}. The initial node features $\mathbf{X}$ are transformed to learned embeddings $\mathbf{Z}$ through the use of a transformation function $f_T$: 
\begin{equation}
     \mathbf{Z} = f_T(G,\mathbf{X})
     \label{eq:transformation}
\end{equation} 
The transformation function $f_T$ can be multi-layers of graph convolutions (with message passing) for feature smoothing or a simple multi-layer perceptron (without any explicit message passing) for dimensionality reduction. We have considered three different types of transformation layers: Nested Network convolution \cite{song2021network}, GCN convolution \cite{kipf2017semi}, and Multi-Layer Perceptron (MLP). Nested Network convolution and GCN convolution are message-passing layers whereas MLP is a graph-agnostic transformation method (see more details in Section \ref{transformer}). After obtaining node embeddings $\mathbf{Z}$, our proposed \poolssnpT function is used to aggregate the target subgraph's internal and external topological properties into a latent subgraph representation:
\begin{equation}
     \mathbf{q}_s = \poolssnp(\mathbf{Z}, G, S)
     \label{eq:readout}
\end{equation} 
This subgraph representation $\mathbf{q}_s$ is fed to an MLP to output class probabilities for the subgraph classification task. The MLP, in addition to giving the class probabilities, learns how to mix the pooled subgraph and its neighborhood representations. Our proposed solution does not require computationally-expensive labeling tricks (as opposed to GLASS \cite{wang2021glass}), or artificially-crafted message passing channels (as opposed to SubGNN \cite{alsentzer2020subgraph}). This computational reduction is achieved by applying transformation on the base graph (rather than on subgraphs) and our proposed SSNP function. Detailed information on transformation layers and our proposed \poolssnpT function follows.





\subsection{Transformation Layer} \label{transformer}
In addition to deploying MLP as a transformation function, we have considered two graph convolution layers. We discuss their formulations in this section. 
\vskip 1mm
\noindent \textbf{Nested Network Convolution.} Our Nested Network (NN) convolution follows a Network in Network architecture \cite{song2021network} as a way of deepening a GNN model by adding multiple non-linear layers within a convolution layer to increase model capacity while preventing overfitting and oversmoothing. The first step of the NN convolution layer is to transform the current layer's node embeddings $\mathbf{h}_u^{(l-1)}$ using one linear layer with an activation function $\sigma$:
\begin{equation}
    \begin{split}
     \mathbf{\hat{h}}_u^{(1)} = \sigma\left(\mathbf{W}^{(l)}_1\mathbf{h}_u^{(l-1)}\right)  \\
     \end{split}
     \label{eq:initial_linearcomgraph}
\end{equation} 
Following this, we perform simple message passing with summation aggregation followed by graph normalization and dropout: 
\begin{equation} 
    \mathbf{\hat{h}}_u^{(2)}=f_{GD}\left(\sum_{v \in N^+(u)}\mathbf{\hat{h}}^{(1)}_v\right)
    \label{eq:multiplicationwithxcomgraph}
\end{equation} 
where $N^+(u)$ contains the neighbors of $u$ and the node $u$ itself and $f_{GD}$ is a sequential function of graph normalization followed by dropout. The recently updated representation $\mathbf{\hat{h}}_u^{(2)}$ is then concatenated with the original layer's input representation $\mathbf{h}_u^{(l-1)}$ (similar to residual connections \cite{resnet,pmlr-v80-xu18c}) to be linearly transformed to the output representation of the layer: 
\begin{equation}
    \mathbf{h}_u^{(l)} = \mathbf{W}_2^{(l)}\left(\mathbf{\hat{h}}_u^{(2)} \oplus \mathbf{h}_u^{(l-1)}\right)
    \label{eq:finalcom}
\end{equation}

Equations \ref{eq:initial_linearcomgraph}, \ref{eq:multiplicationwithxcomgraph} and \ref{eq:finalcom} constitute a single layer of convolution in our NN model with two learnable weight matrices $\mathbf{W}_1^{(l)}$ and $\mathbf{W}_2^{(l)}$. 

\vskip 1mm
\noindent \textbf{GCN Convolution:} Our implemented GCN convolution layers exactly follows GCN \cite{kipf2017semi}. Neighborhood features are aggregated through message-passing by

\begin{equation}
    \begin{split}
     \mathbf{h}_u^{(l)} = \sigma\left(\mathbf{W}^{(l)}\sum_{v \in N^+(u)}\mathbf{h}_v^{(l-1)}\right), \\
     \end{split}
     \label{eq:gcn-conv}
\end{equation} 
where $\mathbf{W}^{(l)}$ is a learnable weight matrix, and $N^+(u)$ contains the neighbors of $u$ and itself.  


\subsection{Subgraph Neighborhood Pooling and Variants}


Our proposed pooling is built based on the idea that the representations of subgraphs and their neighborhoods are both important for capturing the internal and external topology of subgraphs. We first define the $h$-hop subgraph neighborhood as:

\begin{defn}[$h$-hop Subgraph Neighborhood]
Given the base graph $G=(V, E)$ and its subgraph $S=(V_S, E_S)$, the $h$-hop subgraph neighborhood $N_S^{(h)}$ is the induced subgraph created from the node set $\{j \in V_N |  min_{i \in S} d(i,j) \leq h\}$, where $d(i,j)$ is the geodesic distance between node $i$ and $j$, and $V_{N} = V \setminus V_S$ are nodes of $G$ that do not belong to $S$. 
\label{def:complement1}
\end{defn}

\begin{figure}
  \centering
  \includegraphics[width=0.7\linewidth]{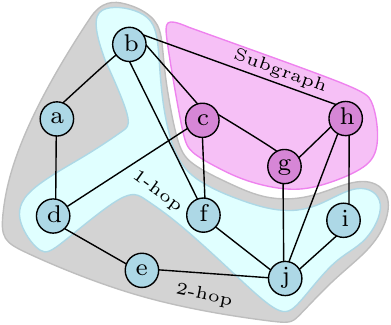}
  \caption{$h$-hop subgraph neighborhood.}
  \label{fig:h-hop}
\end{figure}
In simple words, the $h$-hop subgraph neighborhood is the subgraph of $G$ whose nodes do not belong to $S$ and are within a distance of $h$ to at least one of the nodes of $S$. An example of $1$-hop and $2$-hop subgraph neighborhood is shown in Figure \ref{fig:h-hop}. Our $h$-hop subgraph neighborhood can be viewed as an extension (or generalization) of the enclosing subgraphs for pair of nodes \cite{zhang2018link} but with two distinctions: (i) the $h$-hop neighborhood is defined for any subgraph size (rather than just a pair of nodes) and (ii) the subgraph $S$ is excluded from its neighborhood subgraph. 
Given this $h$-hop subgraph neighborhood definition, we first consider a simple \textit{subgraph neighborhood pooling}:
\begin{equation}
    \poolsnp(\mathbf{Z},G,S, h) = \pool_s(\mathbf{Z}_S,S) \oplus \pool_n\left(\mathbf{Z}_N, N^{(h)}_S\right), 
    \label{eq:snp}
\end{equation}
where $\mathbf{Z}_S$ and $\mathbf{Z}_N$  denote the matrix node embeddings of the subgraph $S$ and its neighborhood $N^{(h)}_S$. Here, $\oplus$ is the concatenation operator, and $pool_s$ and $pool_n$ can be any order invariant graph pooling function (e.g., sum, mean, max, size, or SortPooling \cite{zhang2018end}). The main idea here is simple: treat the subgraph and its neighborhood as two separate graphs, then pool their information, and then concatenate their representations to capture both the internal and external topology of the subgraph. Current subgraph representation learning models (e.g., GLASS, SubGNN) only use $pool_s$, while ignoring the rich information of the neighborhood subgraph. 

\begin{figure*}[t]
\centering
\begin{subfigure}{0.46\linewidth}
\includegraphics[width=\linewidth]{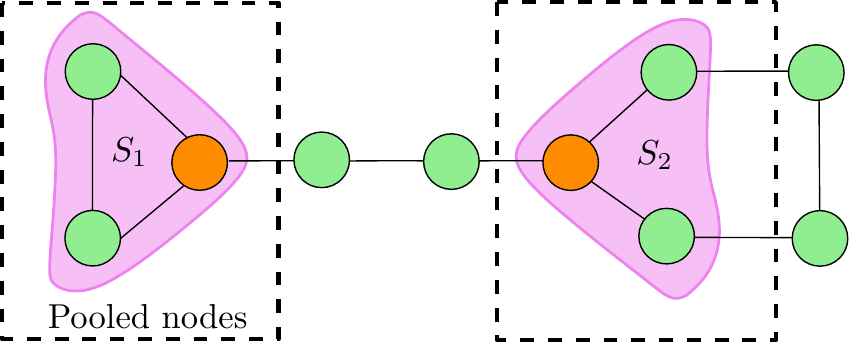}
\caption{Subgraph Pooling, Iter. 1}
\label{fig:subfig1}
\end{subfigure}\hfil
\begin{subfigure}{0.46\linewidth}
\includegraphics[width=\linewidth]{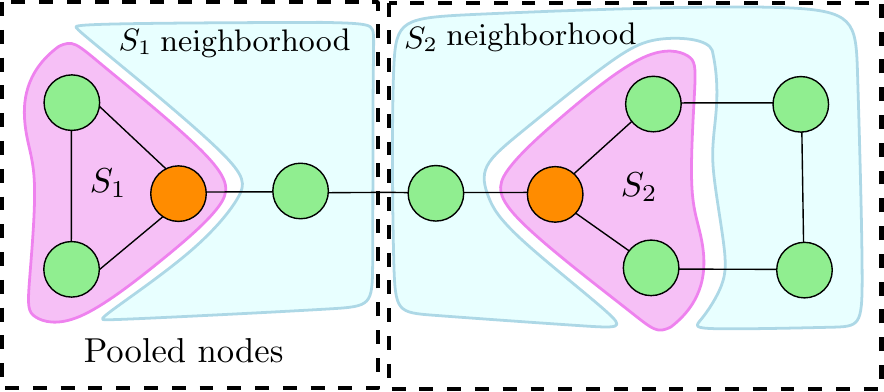}
\caption{Subgraph Neighborhood Pooling, Iter. 1}
\label{fig:subfig2}
    \end{subfigure}

\smallskip
     \begin{subfigure}{0.46\linewidth}
\includegraphics[width=\linewidth]{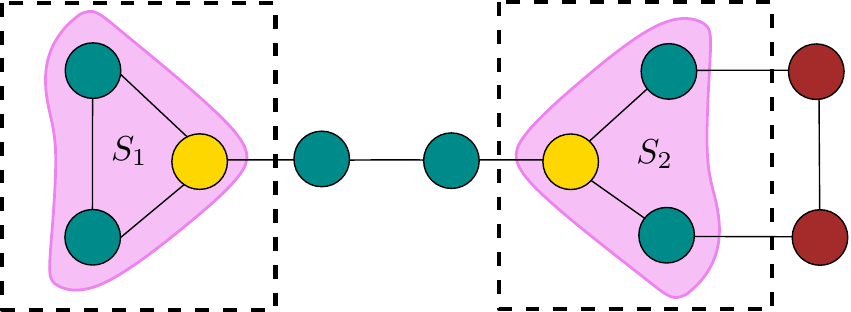}
\caption{Subgraph Pooling, Iter. 2}
\label{fig:subfig3}
    \end{subfigure}\hfil
     \begin{subfigure}{0.46\linewidth}
\includegraphics[width=\linewidth]{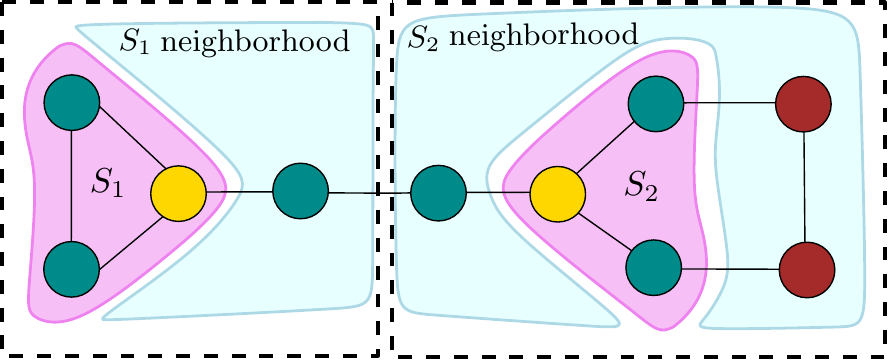}
\caption{Subgraph Neighborhood Pooling, Iter. 2}
\label{fig:subfig4}
    \end{subfigure}

\caption{Comparison of subgraph pooling vs subgraph neighborhood pooling for MPGNNs on distinguishing two non-isomorphic subgraphs $S_1$ and $S_2$ without any distinguishing node features. (a) After one iteration of 1-WL coloring/MPGNN followed by subgraph pooling (shown by pink shaded area), $S_1$ and $S_2$ has the same representation (the nodes involved in the pooling step are in dotted boxes). (b) After one iteration of 1-WL  followed by subgraph neighborhood pooling including both subgraph pooling (shown by pink shaded area) and neighborhood pooling (shown by blue shaded area), $S_1$ and $S_2$ have different representations. (c) After two 1-WL iterations followed by subgraph pooling, $S_1$ and $S_2$ still are not distinguishable. (d) After two 1-WL iterations followed by subgraph neighborhood pooling, $S_1$ and $S_2$ have different representations.}
\label{fig:comparison}
\end{figure*}

However, consuming the complete subgraph neighborhoods is computationally problematic as the subgraph neighborhoods can become extremely large and dense with many uninformative and noisy nodes, thus hindering the model's learning capability and slowing down the running time. 
To overcome this limitation, we define $h$-hop sparsified subgraph neighborhood:

\begin{defn}[$h$-hop Sparsified Subgraph Neighborhood]
Given the base graph $G=(V, E)$ and subgraph $S=(V_S, E_S)$, we define the $h$-hop sparsified subgraph neighborhood $\hat{N}^{(h,k)}_S$, as the subgraph induced from the nodes in $\hat{V}^{(h,k)}_{S} \in \{W_S^{(h,k)} \setminus V_S \}$, where $W_S^{(h,k)}$ is the set of nodes visited by $k$ many $h$-length random-walk(s) from the nodes in $V_{S}$. 
\label{def:complement2}
\end{defn}

Compared to the exact subgraph neighborhood which can get extremely large, the size of the sparse subgraph neighborhoods is bounded by $hk$, which is the product of the length and number of random walks.  The rooted random walks allow sampling ``important" external nodes to a subgraph (similar to rooted PageRank \cite{rootedpr}), which encapsulates information on the border structure and neighborhood. The randomness in the neighborhood subgraph also adds some regularization effect to the training of the model (similar to what was observed in ScaLed \cite{louis2022sampling}). Our $h$-hop sparsified subgraph neighborhood has a resemblance with random-walk sampled enclosing subgraphs \cite{louis2022sampling}, but differs in two ways: the neighborhood does not include the original subgraph, and the neighborhood is defined over arbitrary-sized subgraphs (rather than pair of nodes). Given the computational and learning advantages of specified neighborhood subgraphs, we introduce \textit{stochastic subgraph neighborhood pooling (SSNP)} by a slight modification of Eq. \ref{eq:snp}:  
\begin{equation}
    \poolssnp(\mathbf{Z},G,S,h,k) = \pool_s(\mathbf{Z}_S,S) \oplus \pool_n\left(\mathbf{Z}_N, \hat{N}^{(h,k)}_S\right), 
\end{equation}
where $\mathbf{Z}_S$ and $\mathbf{Z}_N$  denote the matrix node embeddings of the subgraph $S$ and its sparsified neighborhood $\hat{N}^{(h,k)}_S$ by $k$-many $h$-length random walks. In the absence of distinguishing node features, our model with \poolssnpT is more expressive than a plain GNN (which only pools subgraph embeddings without its neighbors). Figure \ref{fig:comparison} shows an example of two subgraphs that are distinguishable under our model, but not under the plain GNN. This additional expressiveness is just an outcome of simple low-cost neighborhood pooling.

Random walks are effective in approximating and sparsifying subgraphs around a node \cite{louis2022sampling,yin2022algorithm}.  However, the sampling of the sparsified subgraph neighborhood in each training epoch might introduce undesirable instability and stochasticity in gradient computations and optimization procedures. To account for this instability as well as manage the sampling overhead, we introduce and distinguish three different stochastic subgraph neighborhood sampling strategies. 
\vskip 1mm
\noindent \textbf{Online Stochastic Views (OV):} The $h$-hop sparsified subgraph neighborhood is sampled in each epoch. This stochasticity over training intends to add implicit regularization to the model but might have undesirable outcomes of gradient instability. Also, the epoch-level sampling adds computational overhead to the training. This computational overhead is due to sampling potentially redundant sparsified subgraph neighborhoods as many times as the total number of epochs.  
\vskip 1mm
\noindent \textbf{Pre-processed Stochastic Views (PV):} To overcome the additional overhead created by sampling during training, we propose \textit{pre-processed stochastic views (PV)} for which a fixed number $n_v$ of sparsified subgraph neighborhood is sampled for each subgraph before training (i.e., during preprocessing). These sampled neighborhood subgraphs can be viewed as data augmentation that provides $n_v$ views of the subgraph neighborhood. Similar to other data augmentation strategies, PV improves the generalization of our model and makes it more robust to noise and overfitting.\footnote{The impact of multi-view augmentations on subgraphs has also been studied recently \cite{shen2022improving,liu2023position}. However, our augmentation techniques create multiple views of the subgraph neighborhoods rather than subgraphs.}. However, the dataset size and training time grows linearly with the number of views $n_v$. 
\vskip 1mm
\noindent \textbf{Pre-processed Online Stochastic Views (POV):} To reduce the training time on the augmented datasets, we propose \textit{pre-processed online stochastic views (POV)}  that leverages both the pre-processed and online subgraph neighborhood sampling method. In the pre-processing stage similar to PV, POV creates $n_v$ multiple sparsified subgraph neighborhoods (i.e., multiple views) for each subgraph. But, during each epoch of training, for each subgraph only $n_{ve}$ of the precomputed views are randomly sampled for training. POV allows data augmentation with multiple views while keeping the number of training instances per epoch independent of the number of views $n_v$.  To do so, we have introduced the number of views per epoch $n_{ve}$.\footnote{Unlike contrastive learning methods, our model does not jointly learn from the different subgraph neighborhood views. As a result, our model is much faster than contrastive learning models.} 

\begin{table}
\centering
\scalebox{0.8}{
\begin{tabular}{lccccc} 
\toprule
          & \textbf{\# nodes} & \textbf{\# edges} & \textbf{\# Subgraphs} & \textbf{\# Classes} & \textbf{Multi-label}  \\ 
\midrule
\textrm{ppi-bp}    & 17080                         & 316951                        & 1591                                     & 6                                      & No                     \\
\textrm{hpo-metab} & 14587                         & 3238174                      & 2400                                     & 6                                      & No                     \\
\textrm{hpo-neuro} & 14587                         & 3238174                      & 4000                                     & 10                                     & Yes                     \\
\textrm{em-user}   & 57333                         & 4573417                      & 324                                       & 2                                      & No                     \\
\bottomrule
\end{tabular}}
\caption{Statistics of all real-world datasets.}
\label{table:statistics}
\end{table}

\begin{table*}[t!]
\centering
\begin{tabular}{p{2.1cm}>{\centering}p{2.4cm}>{\centering}p{2.4cm}>{\centering}p{2.4cm}>{\centering\arraybackslash}p{2.4cm}} 
\toprule
\textbf{Model}                     & \textbf{ppi-bp} & \textbf{hpo-metab} & \textbf{hpo-neuro} & \textbf{em-user}  \\ 
\cmidrule[0.75pt](r){1-1} \cmidrule[0.75pt](lr){2-5}
MLP                       & 0.445$\pm$0.003   & 0.386$\pm$0.011      & 0.404$\pm$0.006      & 0.524$\pm$0.019     \\
GBDT                      & 0.446$\pm$0.000   & 0.404$\pm$0.000      & 0.513$\pm$0.000      & 0.694$\pm$0.000     \\
GNN-plain                 & \third{0.613$\pm$0.009}   & \second{0.597$\pm$0.012}      & 0.668$\pm$0.007      & 0.847$\pm$0.021     \\
Sub2Vec                   & 0.388$\pm$0.001   & 0.472$\pm$0.010      & 0.618$\pm$0.003      & 0.779$\pm$0.013     \\
GNN-seg                   & 0.361$\pm$0.008   & 0.542$\pm$0.009      & 0.647$\pm$0.001      & 0.725$\pm$0.003     \\
SubGNN                    & 0.599$\pm$0.008   & 0.537$\pm$0.008      & 0.644$\pm$0.006      & 0.816$\pm$0.013     \\
GLASS                     & \second{0.618$\pm$0.006}   & \first{0.598$\pm$0.014}      & \second{0.675$\pm$0.007}      & \second{0.884$\pm$0.008}     \\
\midrule
\ssnp-MLP      & 0.591$\pm$0.006   & 0.571$\pm$0.006      & \third{0.669$\pm$0.004}      & \third{0.853$\pm$0.012}     \\
\ssnp-GCN & 0.607$\pm$0.005   & 0.553$\pm$0.011      & 0.667$\pm$0.003      & 0.843$\pm$0.014     \\
\ssnp-NN & \first{0.636$\pm$0.007}   & \third{0.587$\pm$0.010}      & \first{0.682$\pm$0.004}      & \first{0.888$\pm$0.005}     \\
\bottomrule
\end{tabular}
\caption{The mean micro-F1 scores (average of 10 runs) with standard error for all models. The top 3 models are indicated by \first{First}, \second{Second}, and \third{Third}.}
\label{table:f1-results}
\end{table*}



\section{Experiments}
We compare our model with \ssnp and its variants against different subgraph classification baselines on four real-world datasets to evaluate our model in terms of performance and scalability.

\vskip 1mm
\noindent \textbf{Datasets.}
We perform experiments on four publicly-available real-world datasets that have been the main subject of study in other subgraph classification works \cite{wang2021glass,alsentzer2020subgraph}. The dataset statistics are available in Table \ref{table:statistics}. In the \textrm{ppi-bp} dataset, the goal is to predict the cellular function of a group of genes, whereas, in \textrm{hpo-metab} we wish to predict the metabolic disease corresponding to a group of phenotypes. The classification task in \textrm{hpo-neuro} is to predict the neurological disease corresponding to a group of phenotypes. In \textrm{em-user}, we wish to predict the gender of the user given the workout history subgraph.  We follow the same dataset split as GLASS \cite{wang2021glass}:  80/10/10 for train, validation, and test splits.

\vskip 1mm
\noindent \textbf{Baselines.}
We consider the GLASS model \cite{wang2021glass} as our state-of-the-art baseline. Other baselines include SubGNN \cite{alsentzer2020subgraph}, graph-agnostic MLP, and GBDT (gradient-boosted decision trees), GNN-plain, Sub2Vec \cite{adhikari2018sub2vec}, and GNN-seg (learning on segregated subgraphs) \cite{wang2021glass}. All the baseline results, except for GLASS, are taken from \cite{wang2021glass}. The GLASS model is rerun by us to capture the timing values and verify that our setup is identical to the setup of reported results.

\vskip 1mm
\noindent \textbf{Setup.}
For GLASS, we use the best-performing reported hyperparameters to reproduce their results. For our model, we set the transformation layers/functions to either MLP, Nested Network (NN), or Graph Convolution Network (GCN), and the corresponding models are called \ssnp-MLP, \ssnp-NN and \ssnp-GCN, respectively. We use the ELU activation \cite{clevert2015fast} for all transformation layers. We always set the number of walks per node $k = 1$, and let the pooling method for the subgraph and neighborhood be the same (i.e., $\pool_s = \pool_n$). Unless noted otherwise, we use the POV for creating subgraph neighborhood views, where we set the number of views $n_v=20$ and the number of views per epoch $n_{ve}=5$. The other hyperparameters are searched over validation datasets to maximize micro-F1 scores. The search spaces are $\pool_s \in \{sum, size\}$, length of walks $h \in \{1, 5\}$, and the number of transformation layers $\in \{1, 2, 3\}$.  Similar to GLASS, we set the learning rate for ppi-bp to $0.0005$ and hpo-neuro to $0.002$ whereas, for both hpo-metab and em-user, we set it to $0.001$. Our model, similar to GLASS and SubGNN, uses pre-trained 64-dimensional nodal features as the initial features for all datasets. We use Adam optimizer \cite{kingma2014adam} paired with ReduceLROnPlateau learning rate scheduler, which reduces the learning rate on plateauing validation dataset loss values. We set dropout \cite{srivastava2014dropout} to $0.5$ for all models. We use a single-layer MLP to output the class probabilities and always use the cross-entropy loss in our model. Our models with NN and GCN transformation layers are trained for a maximum of 300 epochs in each run with a warm-up of 50 epochs for ppi-bp, hpo-metab and hpo-neuro and warm-up of 10 epochs for em-user. We set patience to 50 epochs for hpo-metab and hpo-neuro and 20 for em-user. Our models with the MLP transformation layer are run for 100 epochs. Our model is implemented in PyTorch Geometric \cite{Fey/Lenssen/2019} and PyTorch \cite{paszke2019pytorch}.\footnote{Our code is available at \href{https://github.com/shweta-jacob/SSNP}{https://github.com/shweta-jacob/SSNP}. We run our experiments on servers with 50 CPUs, 377GB RAM, and 11GB GPUs.} Our results are reported with an average F1-score over 10 runs with different random seeds.

\begin{table*}[t]
\centering
\scalebox{1}{
\begin{tabular}{lcccccccc} 
\toprule
         & \multicolumn{4}{c}{\textbf{ppi-bp}}                                             & \multicolumn{4}{c}{\textbf{hpo-metab}}                                           \\ 
\cmidrule[0.6pt](lr){2-5} \cmidrule[0.6pt](lr){6-9}
\textbf{Model}    & Preproc.                  & Training                   & Inference & Runtime               & Preproc.                  & Training                   & Inference   & Runtime             \\ 
\cmidrule[0.6pt](r){1-1}\cmidrule[0.6pt](lr){2-5} \cmidrule[0.6pt](lr){6-9}
\ssnp-NN  & \textit{8.94$\pm$0.54}                 & 0.38$\pm$0.02                  & 0.02$\pm$0.00    &    129.35$\pm$3.27       & 25.20$\pm$0.84              & 0.73$\pm$0.02                  & 0.05$\pm$0.001        &   159.56$\pm$18.86    \\
\ssnp-GCN & 8.89$\pm$0.71                 & \textit{0.42$\pm$0.02}                  & \textit{0.03$\pm$0.00}     &   \textit{142.38$\pm$3.85}       & \textit{26.13$\pm$1.53}                & \textit{0.94$\pm$0.03}                  & \textit{0.06$\pm$0.00}    &    \textit{209.20$\pm$43.15}       \\
\ssnp-MLP & \textbf{8.79$\pm$0.63}        & \textbf{0.06$\pm$0.02}         & \textbf{0.00$\pm$0.00}     & \textbf{16.00$\pm$0.94} & \textbf{24.81$\pm$0.75}       & \textbf{0.10$\pm$0.02}         & \textbf{0.00$\pm$0.00}     & \textbf{35.00$\pm$1.72}\\
GLASS    & 3.93$\pm$0.10        & 0.78$\pm$0.02        & 0.05$\pm$0.00   & 207.99$\pm$24.76  & 15.99$\pm$0.88       & 2.15$\pm$0.03         & 0.13$\pm$0.00      & 239.48$\pm$33.22\\ 
\midrule
\textbf{Speedup}  & \multicolumn{1}{c}{0.44/0.45} & \multicolumn{1}{c}{1.86/13} & \multicolumn{1}{c}{1.67/25} & \multicolumn{1}{c}{1.46/13} & \multicolumn{1}{c}{0.61/0.64} & \multicolumn{1}{c}{2.29/21.5}  & \multicolumn{1}{c}{2.17/43.33}  & \multicolumn{1}{c}{1.14/6.84}\\ 
\bottomrule
\toprule
         & \multicolumn{4}{c}{\textbf{hpo-neuro}}                                          & \multicolumn{4}{c}{\textbf{em-user}}                                             \\ 
\cmidrule[0.6pt](lr){2-5} \cmidrule[0.6pt](lr){6-9}
\textbf{Model}    & Preproc.                  & Training                   & Inference           & Runtime     & Preproc.                  & Training                   & Inference       & Runtime         \\ 
\cmidrule[0.6pt](r){1-1}\cmidrule[0.6pt](lr){2-5} \cmidrule[0.6pt](lr){6-9}
\ssnp-NN  & \textit{29.67$\pm$1.54}                & 1.27$\pm$0.03                  & 0.05$\pm$0.00    &      202.28$\pm$26.01      & \textit{27.93$\pm$1.41}               & \textit{3.00$\pm$0.04}                  & \textit{0.08$\pm$0.00}          &     \textit{156.81$\pm$32.10}  \\
\ssnp-GCN & \textbf{28.14$\pm$0.81}                & \textit{1.58$\pm$0.05}                  & \textit{0.06$\pm$0.00}         &   \textit{344.14$\pm$44.14}   & 27.62$\pm$0.91                & 1.61$\pm$0.04                  & \textit{0.08$\pm$0.00}      &      108.30$\pm$18.62   \\
\ssnp-MLP & 28.37$\pm$1.13       & \textbf{0.21$\pm$0.01}         & \textbf{0.01$\pm$0.00}    & \textbf{50.00$\pm$1.05} & \textbf{27.52$\pm$1.54}       & \textbf{0.16$\pm$0.01}         & \textbf{0.00$\pm$0.00}     &  \textbf{44.00$\pm$1.71} \\
GLASS    & 16.56$\pm$0.84       & 4.20$\pm$0.04         & 0.25$\pm$0.00   &  511.54$\pm$94.40  & 25.11$\pm$1.61       & 4.93$\pm$0.04         & 0.56$\pm$0.00    &  212.28$\pm$23.51 \\ 
\midrule
\textbf{Speedup}  & \multicolumn{1}{c}{0.56/0.59} & \multicolumn{1}{c}{2.66/20}  & \multicolumn{1}{c}{4.17/25} & \multicolumn{1}{c}{1.49/10.23} & \multicolumn{1}{c}{0.90/0.91} & \multicolumn{1}{c}{1.64/30.81} & \multicolumn{1}{c}{7/140} & \multicolumn{1}{c}{1.35/4.82} \\
\bottomrule
\end{tabular}}
\caption{Our model vs GLASS: dataset preparation time, training time per epoch, inference time per epoch, and total runtime in seconds (mean over 10 runs). The total runtime includes preprocessing, training and inference times. The min/max speedup is the ratio of time taken by GLASS to the time of the slowest/fastest \ssnp model (in italics/bold). The runtimes are rounded to two decimal places; but, the speedups are computed from actual runtimes.}
\label{table:timing_comp}
\end{table*}

\vskip 1mm
\noindent \textbf{Results: F1-Score and Runtime.}
Table \ref{table:f1-results} shows the mean micro-F1 results for all datasets. On ppi-bp, hpo-neuro, and em-user, our \ssnp-NN model outperforms all others with a gain of 0.018, 0.011, and 0.004, respectively. For hpo-metab, \ssnp-NN ranks third with a small margin of $0.011$ compared to GLASS ranked first. This relatively low performance could be attributed to the fact that subgraphs in hpo-metab are dense and therefore, do not need external topological information. Surprisingly, both \ssnp-NN and \ssnp-GCN outperform SubGNN across all the datasets. Even, our simplest model \ssnp-MLP (even without message passing) outperforms SubGNN in all datasets except for ppi-bp for which it has a comparable result. \ssnp-MLP also appears to be relatively competitive by being ranked third in hpo-neuro and em-user. All these results indicate that our models with simple transformation layers but the expressive pooling function of \ssnp can easily outperform more complicated and computationally intensive models. 

Our results in Table \ref{table:f1-results} also provide strong evidence in demonstrating how effective neighborhood pooling (and information) is for subgraph classification. The key difference between GNN-plain and \ssnp-NN is the pooling of neighborhood subgraphs in \ssnp-NN as both use NN architecture. Similarly, \ssnp-MLP surpasses MLP by a significant margin too.

The average of dataset preparation time, training time per epoch, inference time per epoch, and total runtime are captured in Table \ref{table:timing_comp}. 
Our models for all datasets require at most twice the preprocessing times of GLASS due to the sampling of multiple views of the neighborhood subgraphs.\footnote{One can easily reduce the preparation time by tweaking the total number of views created for each subgraph.} However, in return, the training and inference times are 1.5-137$\times$ faster depending on the model variations and datasets. Our best-performing \ssnp-NN has a training speedup of 1.5-3.3$\times$ (min. for em-user and max. for hpo-neuro) and an inference speedup of 2.5-7$\times$ (min. for ppi-bp and max. for em-user). Notably, our \ssnp-MLP is the fastest with maximum training and inference (resp.) speedups of 30$\times$ and 140$\times$ (resp.) in em-user. Cross-examining  Tables \ref{table:f1-results} and \ref{table:timing_comp}, we can observe that \ssnp-MLP vs. GLASS has a speedup of 13-140$\times$ (for both training and inference) with a small negative gain of 0.006--0.031 in F1-score. Similarly, we see a runtime speedup of 4.8-13$\times$ (min. for em-user and max. for ppi-bp) with \ssnp-MLP. These results suggest that our simple models outperform all baselines or were comparable while being multiple magnitudes faster than the current state-of-the-art baselines.

\begin{figure*}[t]
\centering
\begin{subfigure}{0.32\linewidth}
\includegraphics[width=\linewidth]{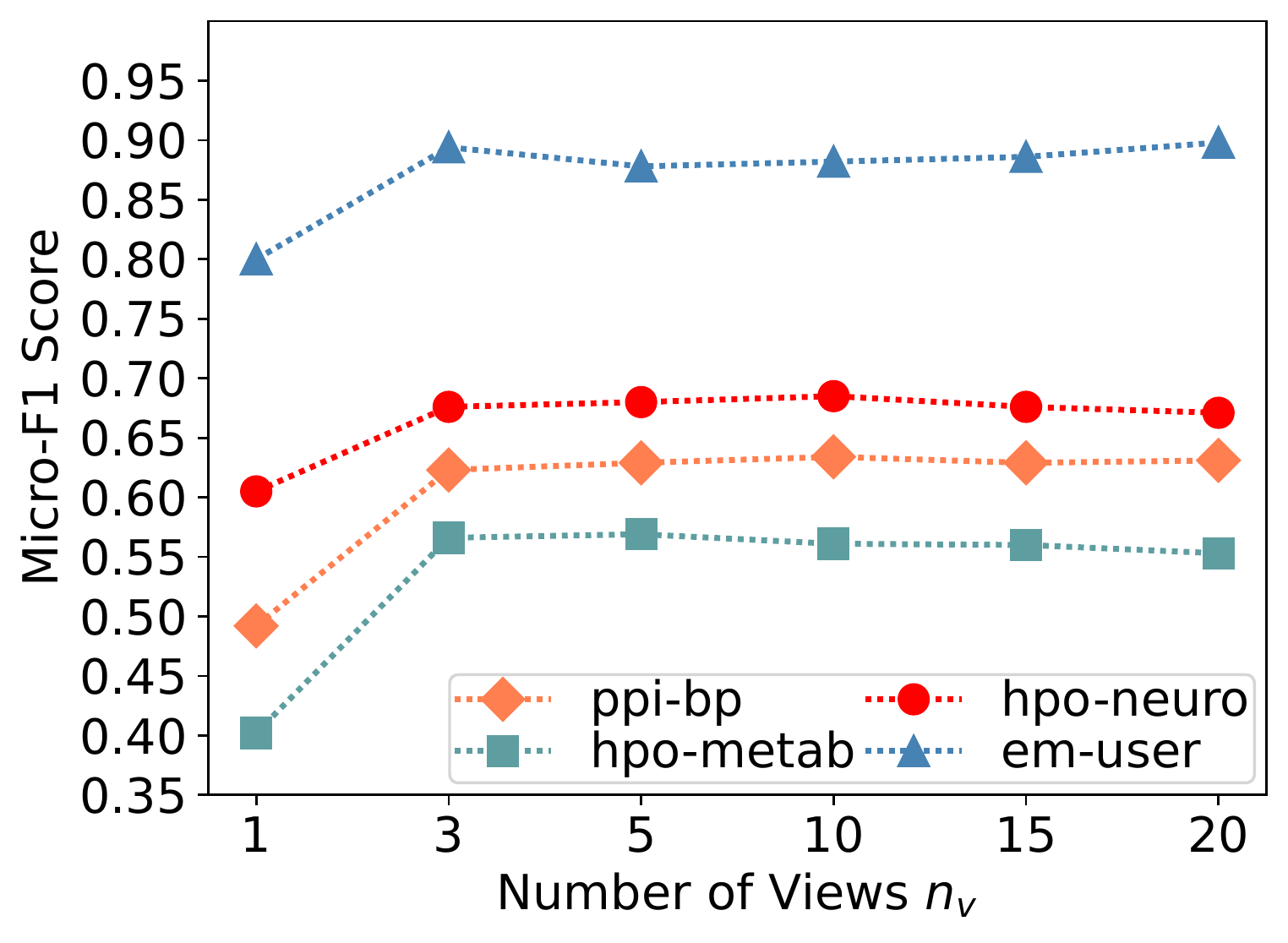}
\caption{\# Views vs F1 Score, PV}
\label{fig:hyperparam_f1_views}
\end{subfigure}\hfil
\begin{subfigure}{0.32\linewidth}
\includegraphics[width=\linewidth]{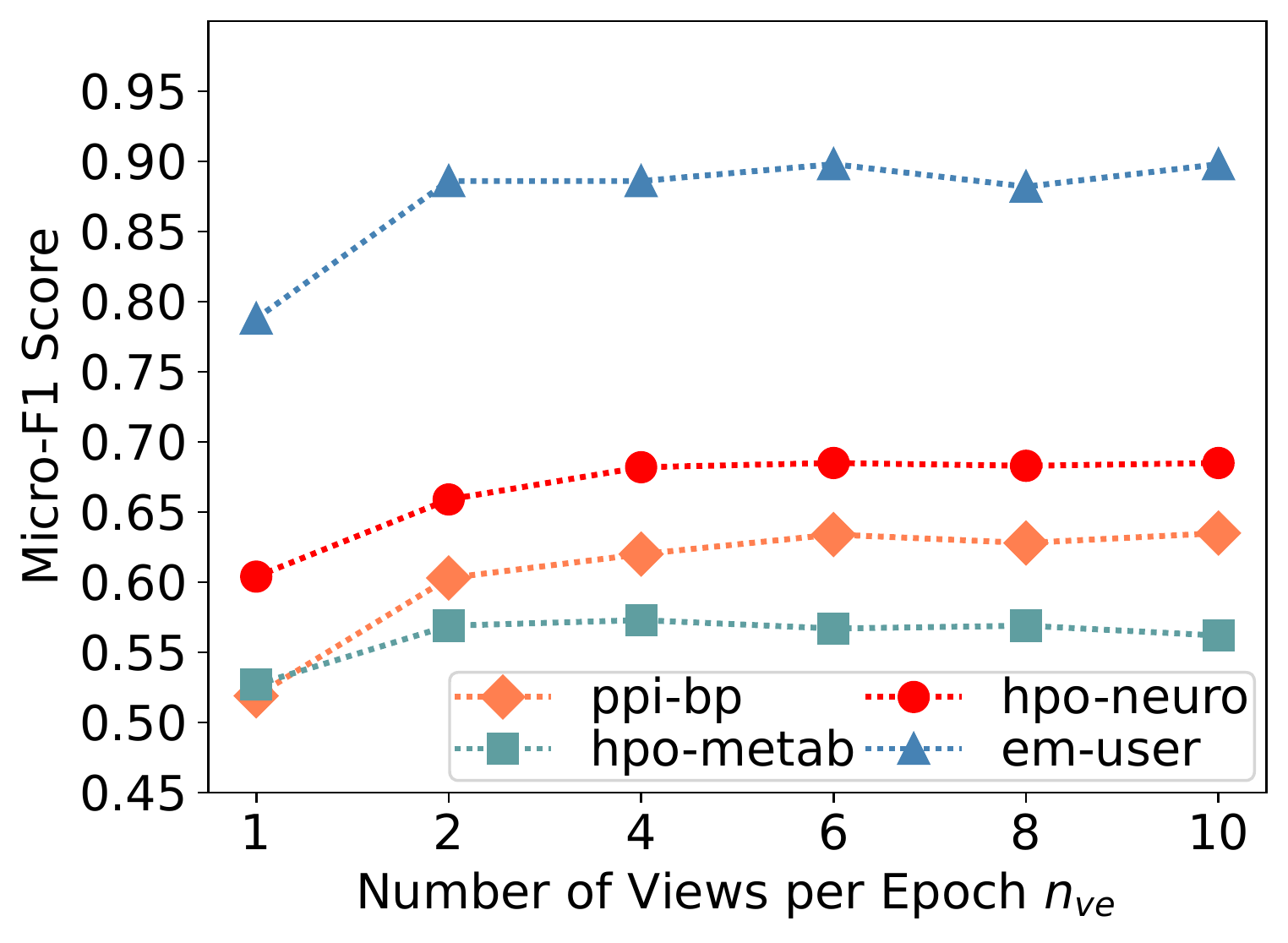}
\caption{\# Views per Epoch vs F1 Score, POV}
\label{fig:hyperparam_f1_views_epoch}
\end{subfigure}
\begin{subfigure}{0.32\linewidth}
\includegraphics[width=\linewidth]{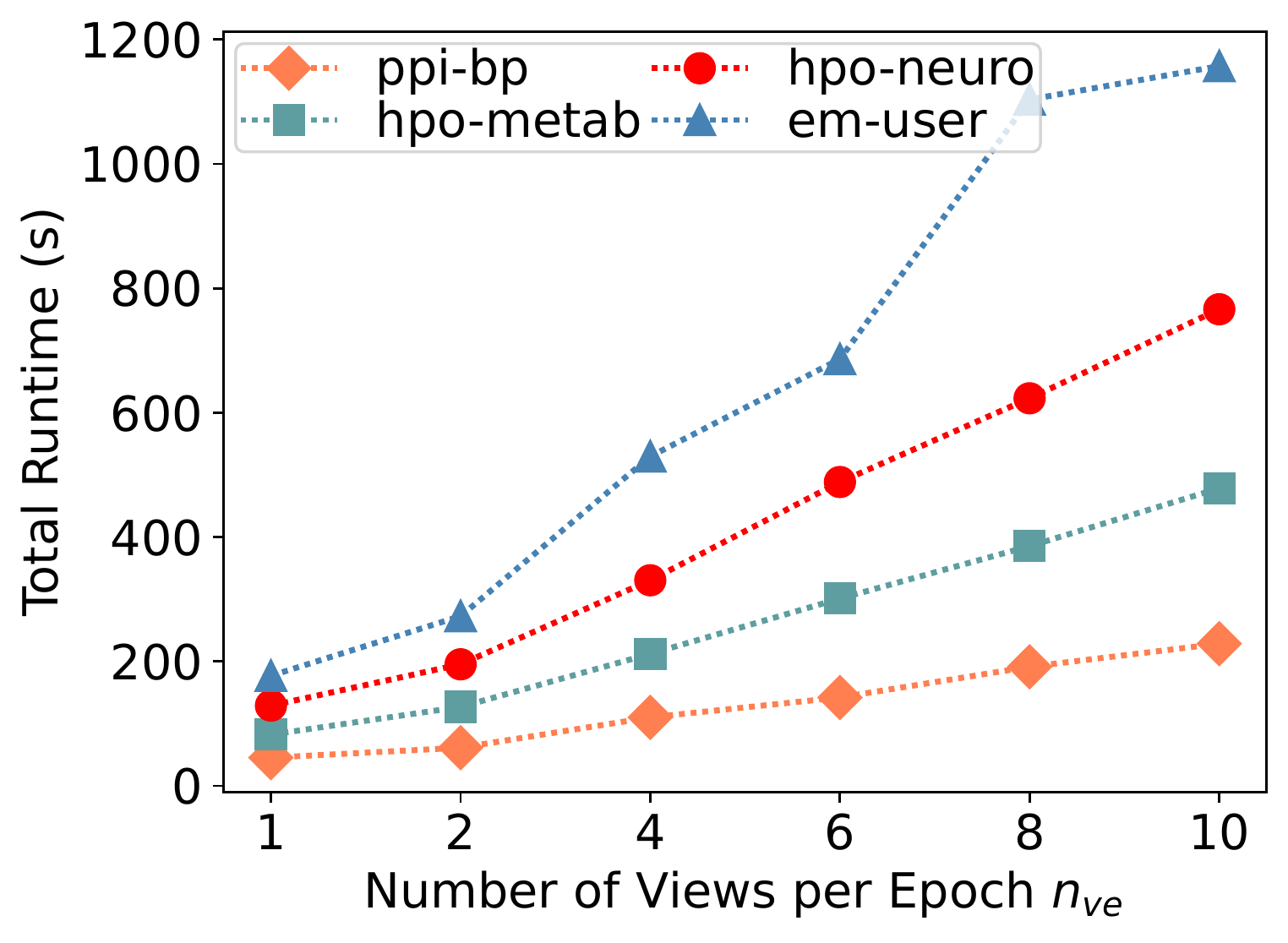}
\caption{\# Views per Epoch vs Runtime, POV}
\label{fig:hyperparam_runtimes_views_epoch}
\end{subfigure}

\caption{Multi-view hyperparameter analyses of \ssnp-NN model variants: (a) The effect of the number of views on F1 score for Pre-processed Stochastic Views (PV). The impact of the number of views per epoch on (b) F1 score and (c) runtime for Pre-processed Online Stochastic views (POV).}
\label{fig:hyperparam}
\end{figure*}

\begin{table*}[t]
\centering
\begin{tabular}{p{3.2cm}>{\centering}p{2.1cm}>{\centering}p{2.1cm}>{\centering}p{2.1cm}>{\centering\arraybackslash}p{2.1cm}}  
\toprule
\textbf{Sampling Strategy} & \textbf{ppi-bp}        & \textbf{hpo-metab}     & \textbf{hpo-neuro}     & \textbf{em-user}        \\ 
\cmidrule[0.75pt](r){1-1} \cmidrule[0.75pt](lr){2-5}
OV                & 0.527$\pm$0.008          & 0.443$\pm$0.055          & 0.681$\pm$0.002          & \textbf{0.906$\pm$0.009}  \\
PV (5 views)      & 0.628$\pm$0.007          & 0.569$\pm$0.015          & 0.680$\pm$0.003          & 0.878$\pm$0.015           \\
PV (20 views)     & 0.635$\pm$0.003          & 0.553$\pm$0.013          & 0.671$\pm$0.003          & 0.902$\pm$0.007           \\
POV               & \textbf{0.638$\pm$0.008} & \textbf{0.577$\pm$0.017} & \textbf{0.686$\pm$0.004} & 0.902$\pm$0.007           \\
\bottomrule
\end{tabular}
\caption{F1-score (avg. over 5 runs) for various sampling strategies, \ssnp-NN.}
\label{table:ablation}
\end{table*}

\begin{figure*}[t]
\centering
\begin{subfigure}{0.365\linewidth}
\includegraphics[width=\linewidth]{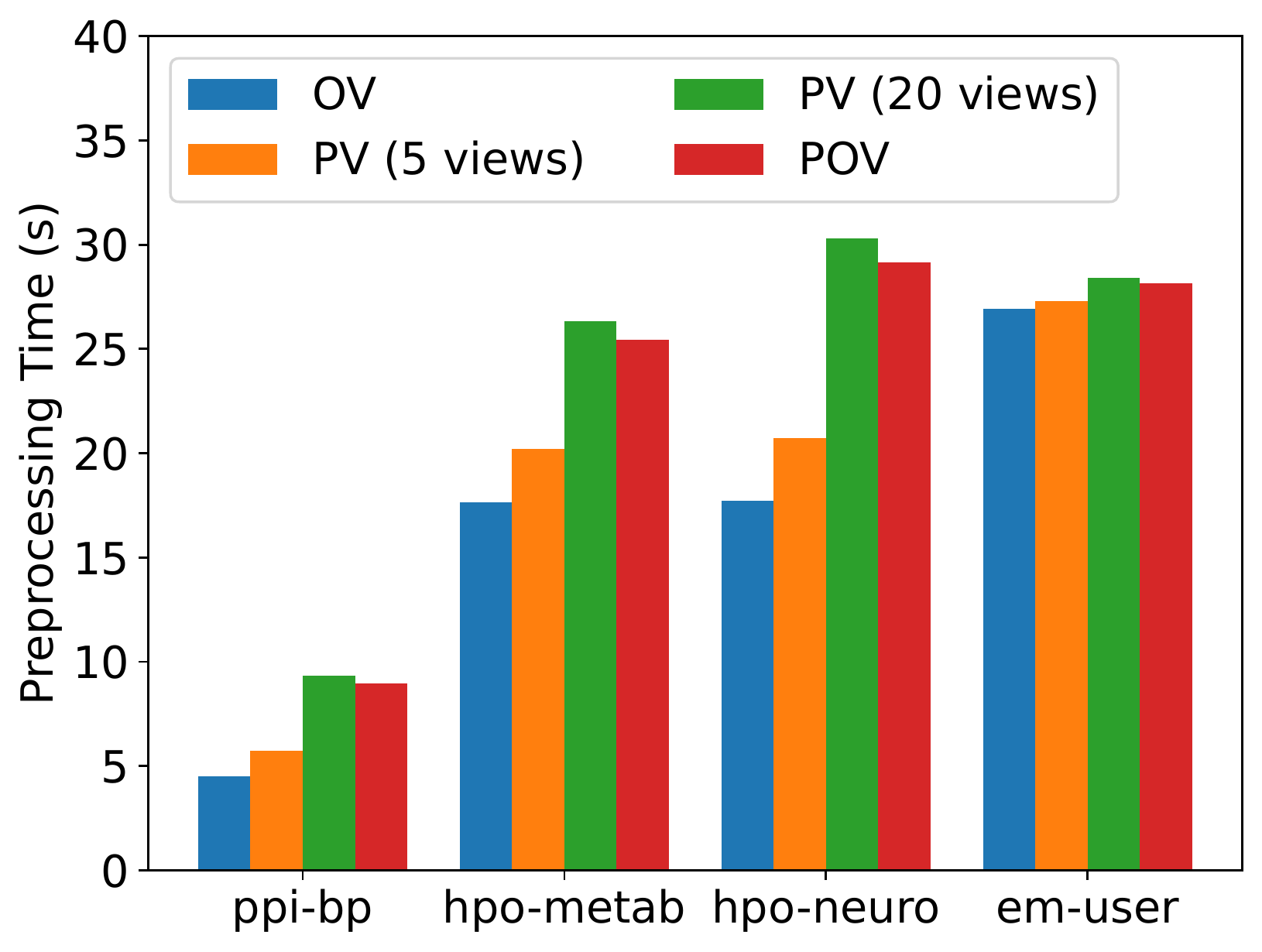}
\label{fig:pre-proc_plot}
\end{subfigure}\hfil
\begin{subfigure}{0.365\linewidth}
\includegraphics[width=\linewidth]{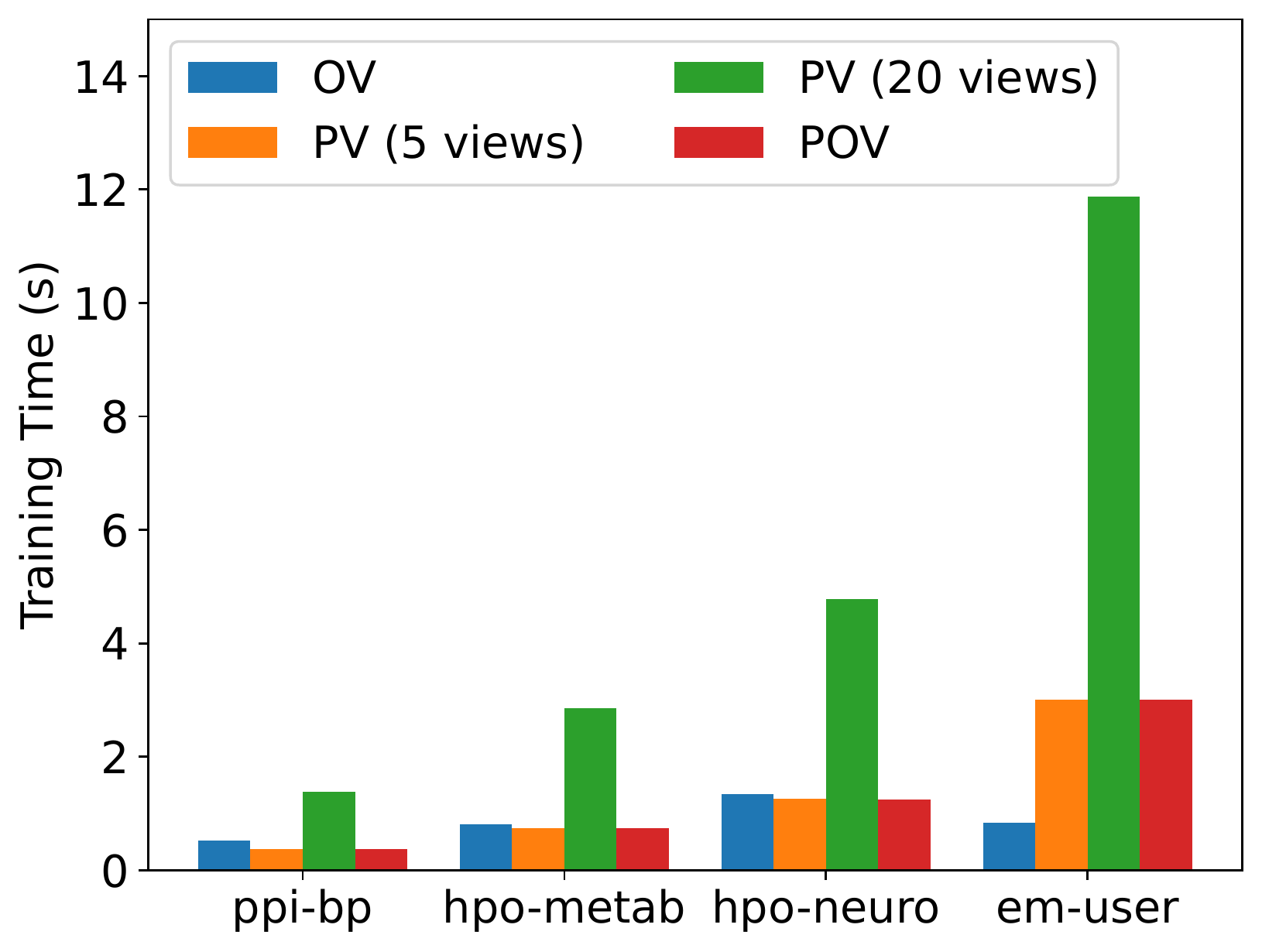}
\label{fig:train_plot}
    \end{subfigure}
    \vspace{-12pt}
    \caption{The effect of sampling strategies on pre-processing time (left) and  training time per epoch (right) in \ssnp-NN.}
    \label{fig:plot}
\end{figure*}

\vskip 1mm
\noindent \textbf{Results: Multi-view Hyperparameter Analyses.}
We first study the effect of the number of views $n_{v}$ in the PV variant of our \ssnp-NN model. For this analysis, we fix $k=1$ and $h=1$ for all datasets (except for em-user and hpo-metab with $h=5$) while changing $n_{v} \in \{1, 3, 5, 10, 15, 20\}$. As shown in Figure \ref{fig:hyperparam_f1_views}, the F1 score for all datasets sharply increases from 1 to 3 and then stabilizes. For hpo-metab, we observe a slight downgrade for a relatively large number of views (e.g., 15 or 20) whereas the F1 score of em-user  achieves its highest score on 20 views. These results suggest that the number of views should be a few (e.g., $n_v=3$ or $n_v=5$), but not so high (e.g, $n_v=20$) to perform consistently over all the datasets.
%
%
%
%
 We further our analyses by studying the effect of the number of views per epoch $n_{ve}$ in the POV variant for a fixed number of views $n_v=10$.  Figure \ref{fig:hyperparam_f1_views_epoch} shows that the F1 score increases with $n_{ve}$, but it has a diminishing return pattern. However, an increase in $n_{ve}$ directly increases the time taken for training and thereby increases the total runtime (see Figure \ref{fig:hyperparam_runtimes_views_epoch}). Surprisingly, $n_{ve}=4$ offers almost the same F1 score as what $n_{ve}=10$ can offer, while requiring considerably less computation time. We believe this performance is primarily due to accessing large enough augmented training data and the regularization offered through the stochasticity of sampled views per epoch. Cross-examination of Figures \ref{fig:hyperparam_f1_views_epoch} and \ref{fig:hyperparam_runtimes_views_epoch} suggest that setting $n_{ve}$ to 2 or 4 offers a good F1-score with manageable computational overhead.

\vskip 1mm
\noindent \textbf{Results: Stochastic Pooling Strategies.}
We intend to study the effect of various stochastic sampling strategies on our \ssnp-NN model. We fix all hyperparameters as was reported above except those related to our pooling strategies. We set the number of views per epoch $n_v$ to 1 for online views (OV), to 5 or 20 for pre-processed views (PV), and to 20 for pre-processed online views (POV). For POV, we also set the number of views per epoch $n_{ve}$ to 5.  

The micro-F1 scores are captured in Table \ref{table:ablation}. The effect of the sampling on the pre-processing and training times are captured in Figure \ref{fig:plot}. For all datasets (except em-user), POV provides the best F1-scores (see Table \ref{table:ablation}). For em-user, OV suppresses POV with a small margin of 0.004. In Figure \ref{fig:plot}, we can see that the average training time for OV in ppi-bp, hpo-metab and hpo-neuro is higher than PV with 5 views and POV. However, pre-processing of OV is faster than all other sampling strategies. For PVs and POV, the pre-processing times are comparable; however, POV offers much faster training time and a higher F1-score (see Table  \ref{table:ablation}). In all datasets (except em-user), the F1 score of PV with 5 views is higher than that of PV with 20 views, implying that a higher number of views does not necessarily improve performance for PV. However, POV, with 5 views per epoch and a total of 20 views, has the highest F1 score. This means that the stochasticity in the views across epochs allows a better generalization for our model. 


\section{Conclusions and Future Work}
The state-of-the-art subgraph classification solutions are not scalable due to the use of labeling tricks or artificial message-passing channels for subgraphs. In this paper, we propose a simple yet powerful model that has our proposed stochastic subgraph neighborhood pooling (\ssnp) in its core. Leveraging \ssnp, our model learns the internal connectivity and border neighborhood of subgraphs. We also present simple data augmentation techniques that help to improve the generalization of our model. Our model combined with our data augmentation techniques outperforms current state-of-the-art subgraph classification models on 3 out of 4 datasets with a speedup of 1.5-3$\times$. For future work, we plan to explore alternative ways to approximate neighborhood subgraphs and combine subgraphs and their neighborhoods during pooling. Another promising direction might be contrastive learning on the different stochastic views of neighborhood subgraphs.

\bibliographystyle{ACM-Reference-Format}
\bibliography{references}


\begin{thebibliography}{47}


\ifx \showCODEN    \undefined \def \showCODEN     #1{\unskip}     \fi
\ifx \showDOI      \undefined \def \showDOI       #1{#1}\fi
\ifx \showISBNx    \undefined \def \showISBNx     #1{\unskip}     \fi
\ifx \showISBNxiii \undefined \def \showISBNxiii  #1{\unskip}     \fi
\ifx \showISSN     \undefined \def \showISSN      #1{\unskip}     \fi
\ifx \showLCCN     \undefined \def \showLCCN      #1{\unskip}     \fi
\ifx \shownote     \undefined \def \shownote      #1{#1}          \fi
\ifx \showarticletitle \undefined \def \showarticletitle #1{#1}   \fi
\ifx \showURL      \undefined \def \showURL       {\relax}        \fi
\providecommand\bibfield[2]{#2}
\providecommand\bibinfo[2]{#2}
\providecommand\natexlab[1]{#1}
\providecommand\showeprint[2][]{arXiv:#2}

\bibitem[Adhikari et~al\mbox{.}(2018)]%
        {adhikari2018sub2vec}
\bibfield{author}{\bibinfo{person}{Bijaya Adhikari}, \bibinfo{person}{Yao
  Zhang}, \bibinfo{person}{Naren Ramakrishnan}, {and} \bibinfo{person}{B~Aditya
  Prakash}.} \bibinfo{year}{2018}\natexlab{}.
\newblock \showarticletitle{Sub2vec: Feature learning for subgraphs}. In
  \bibinfo{booktitle}{\emph{Advances in Knowledge Discovery and Data Mining:
  22nd Pacific-Asia Conference}}. Springer, \bibinfo{pages}{170--182}.
\newblock


\bibitem[Alsentzer et~al\mbox{.}(2020)]%
        {alsentzer2020subgraph}
\bibfield{author}{\bibinfo{person}{Emily Alsentzer}, \bibinfo{person}{Samuel
  Finlayson}, \bibinfo{person}{Michelle Li}, {and} \bibinfo{person}{Marinka
  Zitnik}.} \bibinfo{year}{2020}\natexlab{}.
\newblock \showarticletitle{Subgraph Neural Networks}.
\newblock \bibinfo{journal}{\emph{Advances in Neural Information Processing
  Systems}} (\bibinfo{year}{2020}), \bibinfo{pages}{8017--8029}.
\newblock


\bibitem[Alsentzer et~al\mbox{.}(2022)]%
        {alsentzer2022deep}
\bibfield{author}{\bibinfo{person}{Emily Alsentzer},
  \bibinfo{person}{Michelle~M Li}, \bibinfo{person}{Shilpa~N Kobren},
  \bibinfo{person}{Undiagnosed~Diseases Network}, \bibinfo{person}{Isaac~S
  Kohane}, {and} \bibinfo{person}{Marinka Zitnik}.}
  \bibinfo{year}{2022}\natexlab{}.
\newblock \showarticletitle{Deep learning for diagnosing patients with rare
  genetic diseases}.
\newblock \bibinfo{journal}{\emph{medRxiv}} (\bibinfo{year}{2022}),
  \bibinfo{pages}{2022--12}.
\newblock


\bibitem[Brin and Page(2012)]%
        {rootedpr}
\bibfield{author}{\bibinfo{person}{Sergey Brin} {and} \bibinfo{person}{Lawrence
  Page}.} \bibinfo{year}{2012}\natexlab{}.
\newblock \showarticletitle{Reprint of: The anatomy of a large-scale
  hypertextual web search engine}.
\newblock \bibinfo{journal}{\emph{Computer Networks}} (\bibinfo{year}{2012}),
  \bibinfo{pages}{3825--3833}.
\newblock


\bibitem[Bruna et~al\mbox{.}(2013)]%
        {bruna2013spectral}
\bibfield{author}{\bibinfo{person}{Joan Bruna}, \bibinfo{person}{Wojciech
  Zaremba}, \bibinfo{person}{Arthur Szlam}, {and} \bibinfo{person}{Yann
  LeCun}.} \bibinfo{year}{2013}\natexlab{}.
\newblock \showarticletitle{Spectral Networks and Locally Connected Networks on
  Graphs}.
\newblock \bibinfo{journal}{\emph{arXiv preprint arXiv:1312.6203}}
  (\bibinfo{year}{2013}).
\newblock


\bibitem[Cai and Ji(2020)]%
        {cai2020multi}
\bibfield{author}{\bibinfo{person}{Lei Cai} {and} \bibinfo{person}{Shuiwang
  Ji}.} \bibinfo{year}{2020}\natexlab{}.
\newblock \showarticletitle{A Multi-scale Approach for Graph Link Prediction}.
  In \bibinfo{booktitle}{\emph{Proceedings of the AAAI Conference on Artificial
  Intelligence}}. \bibinfo{pages}{3308--3315}.
\newblock


\bibitem[Chamberlain et~al\mbox{.}(2023)]%
        {chamberlain2023graph}
\bibfield{author}{\bibinfo{person}{Benjamin~Paul Chamberlain},
  \bibinfo{person}{Sergey Shirobokov}, \bibinfo{person}{Emanuele Rossi},
  \bibinfo{person}{Fabrizio Frasca}, \bibinfo{person}{Thomas Markovich},
  \bibinfo{person}{Nils Hammerla}, \bibinfo{person}{Michael~M Bronstein}, {and}
  \bibinfo{person}{Max Hansmire}.} \bibinfo{year}{2023}\natexlab{}.
\newblock \showarticletitle{Graph Neural Networks for Link Prediction with
  Subgraph Sketching}. In \bibinfo{booktitle}{\emph{International Conference on
  Learning Representations}}.
\newblock


\bibitem[Clevert et~al\mbox{.}(2015)]%
        {clevert2015fast}
\bibfield{author}{\bibinfo{person}{Djork-Arn{\'e} Clevert},
  \bibinfo{person}{Thomas Unterthiner}, {and} \bibinfo{person}{Sepp
  Hochreiter}.} \bibinfo{year}{2015}\natexlab{}.
\newblock \showarticletitle{Fast and accurate deep network learning by
  exponential linear units (elus)}.
\newblock \bibinfo{journal}{\emph{arXiv preprint arXiv:1511.07289}}
  (\bibinfo{year}{2015}).
\newblock


\bibitem[Defferrard et~al\mbox{.}(2016)]%
        {defferrard2016convolutional}
\bibfield{author}{\bibinfo{person}{Micha\"{e}l Defferrard},
  \bibinfo{person}{Xavier Bresson}, {and} \bibinfo{person}{Pierre
  Vandergheynst}.} \bibinfo{year}{2016}\natexlab{}.
\newblock \showarticletitle{Convolutional Neural Networks on Graphs with Fast
  Localized Spectral Filtering}. In \bibinfo{booktitle}{\emph{Advances in
  Neural Information Processing Systems}}. \bibinfo{numpages}{9}~pages.
\newblock


\bibitem[Fey and Lenssen(2019)]%
        {Fey/Lenssen/2019}
\bibfield{author}{\bibinfo{person}{Matthias Fey} {and} \bibinfo{person}{Jan~E.
  Lenssen}.} \bibinfo{year}{2019}\natexlab{}.
\newblock \showarticletitle{Fast Graph Representation Learning with {PyTorch
  Geometric}}. In \bibinfo{booktitle}{\emph{ICLR Workshop on Representation
  Learning on Graphs and Manifolds}}.
\newblock


\bibitem[Frasca et~al\mbox{.}(2020)]%
        {rossi2020sign}
\bibfield{author}{\bibinfo{person}{Fabrizio Frasca}, \bibinfo{person}{Emanuele
  Rossi}, \bibinfo{person}{Davide Eynard}, \bibinfo{person}{Benjamin
  Chamberlain}, \bibinfo{person}{Michael Bronstein}, {and}
  \bibinfo{person}{Federico Monti}.} \bibinfo{year}{2020}\natexlab{}.
\newblock \showarticletitle{SIGN: Scalable Inception Graph Neural Networks}. In
  \bibinfo{booktitle}{\emph{ICML 2020 Workshop on Graph Representation Learning
  and Beyond}}.
\newblock


\bibitem[Grover and Leskovec(2016)]%
        {grover2016node2vec}
\bibfield{author}{\bibinfo{person}{Aditya Grover} {and} \bibinfo{person}{Jure
  Leskovec}.} \bibinfo{year}{2016}\natexlab{}.
\newblock \showarticletitle{node2vec: Scalable feature learning for networks}.
  In \bibinfo{booktitle}{\emph{Proceedings of the 22nd ACM SIGKDD international
  conference on Knowledge discovery and data mining}}.
  \bibinfo{pages}{855--864}.
\newblock


\bibitem[Hamilton et~al\mbox{.}(2017)]%
        {hamilton2017inductive}
\bibfield{author}{\bibinfo{person}{William~L. Hamilton}, \bibinfo{person}{Rex
  Ying}, {and} \bibinfo{person}{Jure Leskovec}.}
  \bibinfo{year}{2017}\natexlab{}.
\newblock \showarticletitle{Inductive Representation Learning on Large Graphs}.
  In \bibinfo{booktitle}{\emph{Proceedings of the 31st International Conference
  on Neural Information Processing Systems}}. \bibinfo{pages}{1025–1035}.
\newblock


\bibitem[He et~al\mbox{.}(2016)]%
        {resnet}
\bibfield{author}{\bibinfo{person}{Kaiming He}, \bibinfo{person}{Xiangyu
  Zhang}, \bibinfo{person}{Shaoqing Ren}, {and} \bibinfo{person}{Jian Sun}.}
  \bibinfo{year}{2016}\natexlab{}.
\newblock \showarticletitle{Deep Residual Learning for Image Recognition}. In
  \bibinfo{booktitle}{\emph{2016 IEEE Conference on Computer Vision and Pattern
  Recognition (CVPR)}}. \bibinfo{pages}{770--778}.
\newblock


\bibitem[Huang et~al\mbox{.}(2023)]%
        {huang2023boosting}
\bibfield{author}{\bibinfo{person}{Yinan Huang}, \bibinfo{person}{Xingang
  Peng}, \bibinfo{person}{Jianzhu Ma}, {and} \bibinfo{person}{Muhan Zhang}.}
  \bibinfo{year}{2023}\natexlab{}.
\newblock \showarticletitle{Boosting the Cycle Counting Power of Graph Neural
  Networks with $\text{I} ^2$-GNNs}. In \bibinfo{booktitle}{\emph{The Eleventh
  International Conference on Learning Representations}}.
\newblock


\bibitem[Jiang et~al\mbox{.}(2021)]%
        {jiang2021could}
\bibfield{author}{\bibinfo{person}{Dejun Jiang}, \bibinfo{person}{Zhenxing Wu},
  \bibinfo{person}{Chang-Yu Hsieh}, \bibinfo{person}{Guangyong Chen},
  \bibinfo{person}{Ben Liao}, \bibinfo{person}{Zhe Wang}, \bibinfo{person}{Chao
  Shen}, \bibinfo{person}{Dongsheng Cao}, \bibinfo{person}{Jian Wu}, {and}
  \bibinfo{person}{Tingjun Hou}.} \bibinfo{year}{2021}\natexlab{}.
\newblock \showarticletitle{Could graph neural networks learn better molecular
  representation for drug discovery? A comparison study of descriptor-based and
  graph-based models}.
\newblock \bibinfo{journal}{\emph{Journal of cheminformatics}}
  (\bibinfo{year}{2021}), \bibinfo{pages}{1--23}.
\newblock


\bibitem[Kim and Oh(2022)]%
        {kim2022efficient}
\bibfield{author}{\bibinfo{person}{Dongkwan Kim} {and} \bibinfo{person}{Alice
  Oh}.} \bibinfo{year}{2022}\natexlab{}.
\newblock \showarticletitle{Efficient Representation Learning of Subgraphs by
  Subgraph-To-Node Translation}. In \bibinfo{booktitle}{\emph{ICLR 2022
  Workshop on Geometrical and Topological Representation Learning}}.
\newblock


\bibitem[Kingma and Ba(2015)]%
        {kingma2014adam}
\bibfield{author}{\bibinfo{person}{Diederik~P. Kingma} {and}
  \bibinfo{person}{Jimmy Ba}.} \bibinfo{year}{2015}\natexlab{}.
\newblock \showarticletitle{Adam: {A} Method for Stochastic Optimization}. In
  \bibinfo{booktitle}{\emph{International Conference on Learning
  Representations}}.
\newblock


\bibitem[Kipf and Welling(2017)]%
        {kipf2017semi}
\bibfield{author}{\bibinfo{person}{Thomas~N. Kipf} {and} \bibinfo{person}{Max
  Welling}.} \bibinfo{year}{2017}\natexlab{}.
\newblock \showarticletitle{Semi-Supervised Classification with Graph
  Convolutional Networks}. In \bibinfo{booktitle}{\emph{International
  Conference on Learning Representations}}.
\newblock


\bibitem[Le and Mikolov(2014)]%
        {le2014distributed}
\bibfield{author}{\bibinfo{person}{Quoc Le} {and} \bibinfo{person}{Tomas
  Mikolov}.} \bibinfo{year}{2014}\natexlab{}.
\newblock \showarticletitle{Distributed representations of sentences and
  documents}. In \bibinfo{booktitle}{\emph{International conference on machine
  learning}}. \bibinfo{pages}{1188--1196}.
\newblock


\bibitem[Li et~al\mbox{.}(2021)]%
        {lidistance}
\bibfield{author}{\bibinfo{person}{Boning Li}, \bibinfo{person}{Yingce Xia},
  \bibinfo{person}{Shufang Xie}, \bibinfo{person}{Lijun Wu}, {and}
  \bibinfo{person}{Tao Qin}.} \bibinfo{year}{2021}\natexlab{}.
\newblock \showarticletitle{Distance-enhanced graph neural network for link
  prediction}. In \bibinfo{booktitle}{\emph{ICML 2021 Workshop on Computational
  Biology}}.
\newblock


\bibitem[Liu et~al\mbox{.}(2023)]%
        {liu2023position}
\bibfield{author}{\bibinfo{person}{Chang Liu}, \bibinfo{person}{Yuwen Yang},
  \bibinfo{person}{Zhe Xie}, \bibinfo{person}{Hongtao Lu}, {and}
  \bibinfo{person}{Yue Ding}.} \bibinfo{year}{2023}\natexlab{}.
\newblock \showarticletitle{Position-Aware Subgraph Neural Networks with
  Data-Efficient Learning}. In \bibinfo{booktitle}{\emph{Proceedings of the
  Sixteenth ACM International Conference on Web Search and Data Mining}}.
  \bibinfo{pages}{643--651}.
\newblock


\bibitem[Louis et~al\mbox{.}(2022)]%
        {louis2022sampling}
\bibfield{author}{\bibinfo{person}{Paul Louis}, \bibinfo{person}{Shweta~Ann
  Jacob}, {and} \bibinfo{person}{Amirali Salehi-Abari}.}
  \bibinfo{year}{2022}\natexlab{}.
\newblock \showarticletitle{Sampling Enclosing Subgraphs for Link Prediction}.
  In \bibinfo{booktitle}{\emph{Proceedings of the 31st ACM International
  Conference on Information \& Knowledge Management}}.
  \bibinfo{pages}{4269--4273}.
\newblock


\bibitem[Louis et~al\mbox{.}(2023)]%
        {louis2023simplifying}
\bibfield{author}{\bibinfo{person}{Paul Louis}, \bibinfo{person}{Shweta~Ann
  Jacob}, {and} \bibinfo{person}{Amirali Salehi-Abari}.}
  \bibinfo{year}{2023}\natexlab{}.
\newblock \showarticletitle{Simplifying Subgraph Representation Learning for
  Scalable Link Prediction}.
\newblock \bibinfo{journal}{\emph{arXiv preprint arXiv:2301.12562}}
  (\bibinfo{year}{2023}).
\newblock


\bibitem[Morselli~Gysi et~al\mbox{.}(2021)]%
        {morselli2021network}
\bibfield{author}{\bibinfo{person}{Deisy Morselli~Gysi},
  \bibinfo{person}{{\'I}talo Do~Valle}, \bibinfo{person}{Marinka Zitnik},
  \bibinfo{person}{Asher Ameli}, \bibinfo{person}{Xiao Gan},
  \bibinfo{person}{Onur Varol}, \bibinfo{person}{Susan~Dina Ghiassian},
  \bibinfo{person}{JJ Patten}, \bibinfo{person}{Robert~A Davey},
  \bibinfo{person}{Joseph Loscalzo}, {et~al\mbox{.}}}
  \bibinfo{year}{2021}\natexlab{}.
\newblock \showarticletitle{Network medicine framework for identifying
  drug-repurposing opportunities for COVID-19}.
\newblock \bibinfo{journal}{\emph{Proceedings of the National Academy of
  Sciences}} (\bibinfo{year}{2021}).
\newblock


\bibitem[Namanloo et~al\mbox{.}(2022)]%
        {namanloo2022improving}
\bibfield{author}{\bibinfo{person}{Alireza~A Namanloo}, \bibinfo{person}{Julie
  Thorpe}, {and} \bibinfo{person}{Amirali Salehi-Abari}.}
  \bibinfo{year}{2022}\natexlab{}.
\newblock \showarticletitle{Improving Peer Assessment with Graph Neural
  Networks.}
\newblock \bibinfo{journal}{\emph{International Educational Data Mining
  Society}} (\bibinfo{year}{2022}).
\newblock


\bibitem[Pan et~al\mbox{.}(2022)]%
        {pan2022neural}
\bibfield{author}{\bibinfo{person}{Liming Pan}, \bibinfo{person}{Cheng Shi},
  {and} \bibinfo{person}{Ivan Dokmani{\'c}}.} \bibinfo{year}{2022}\natexlab{}.
\newblock \showarticletitle{Neural Link Prediction with Walk Pooling}. In
  \bibinfo{booktitle}{\emph{International Conference on Learning
  Representations}}.
\newblock


\bibitem[Paszke et~al\mbox{.}(2019)]%
        {paszke2019pytorch}
\bibfield{author}{\bibinfo{person}{Adam Paszke}, \bibinfo{person}{Sam Gross},
  \bibinfo{person}{Francisco Massa}, \bibinfo{person}{Adam Lerer},
  \bibinfo{person}{James Bradbury}, \bibinfo{person}{Gregory Chanan},
  \bibinfo{person}{Trevor Killeen}, \bibinfo{person}{Zeming Lin},
  \bibinfo{person}{Natalia Gimelshein}, \bibinfo{person}{Luca Antiga},
  \bibinfo{person}{Alban Desmaison}, \bibinfo{person}{Andreas K\"{o}pf},
  \bibinfo{person}{Edward Yang}, \bibinfo{person}{Zach DeVito},
  \bibinfo{person}{Martin Raison}, \bibinfo{person}{Alykhan Tejani},
  \bibinfo{person}{Sasank Chilamkurthy}, \bibinfo{person}{Benoit Steiner},
  \bibinfo{person}{Lu Fang}, \bibinfo{person}{Junjie Bai}, {and}
  \bibinfo{person}{Soumith Chintala}.} \bibinfo{year}{2019}\natexlab{}.
\newblock \showarticletitle{PyTorch: An Imperative Style, High-Performance Deep
  Learning Library}. In \bibinfo{booktitle}{\emph{Advances in Neural
  Information Processing Systems}}. Article \bibinfo{articleno}{721},
  \bibinfo{numpages}{12}~pages.
\newblock


\bibitem[Perozzi et~al\mbox{.}(2014)]%
        {perozzi2014deepwalk}
\bibfield{author}{\bibinfo{person}{Bryan Perozzi}, \bibinfo{person}{Rami
  Al-Rfou}, {and} \bibinfo{person}{Steven Skiena}.}
  \bibinfo{year}{2014}\natexlab{}.
\newblock \showarticletitle{Deepwalk: Online learning of social
  representations}. In \bibinfo{booktitle}{\emph{Proceedings of the 20th ACM
  SIGKDD international conference on Knowledge discovery and data mining}}.
  \bibinfo{pages}{701--710}.
\newblock


\bibitem[Salehi-Abari and Boutilier(2015)]%
        {salehi2015preference}
\bibfield{author}{\bibinfo{person}{Amirali Salehi-Abari} {and}
  \bibinfo{person}{Craig Boutilier}.} \bibinfo{year}{2015}\natexlab{}.
\newblock \showarticletitle{Preference-oriented social networks: Group
  recommendation and inference}. In \bibinfo{booktitle}{\emph{Proceedings of
  the 9th ACM Conference on Recommender Systems}}. \bibinfo{pages}{35--42}.
\newblock


\bibitem[Shen et~al\mbox{.}(2022)]%
        {shen2022improving}
\bibfield{author}{\bibinfo{person}{Yili Shen}, \bibinfo{person}{Jiaxu Yan},
  \bibinfo{person}{Cheng-Wei Ju}, \bibinfo{person}{Jun Yi},
  \bibinfo{person}{Zhou Lin}, {and} \bibinfo{person}{Hui Guan}.}
  \bibinfo{year}{2022}\natexlab{}.
\newblock \showarticletitle{Improving Subgraph Representation Learning via
  Multi-View Augmentation}.
\newblock \bibinfo{journal}{\emph{arXiv preprint arXiv:2205.13038}}
  (\bibinfo{year}{2022}).
\newblock


\bibitem[Song et~al\mbox{.}(2021)]%
        {song2021network}
\bibfield{author}{\bibinfo{person}{Xiang Song}, \bibinfo{person}{Runjie Ma},
  \bibinfo{person}{Jiahang Li}, \bibinfo{person}{Muhan Zhang}, {and}
  \bibinfo{person}{David~Paul Wipf}.} \bibinfo{year}{2021}\natexlab{}.
\newblock \showarticletitle{Network in graph neural network}.
\newblock \bibinfo{journal}{\emph{arXiv preprint arXiv:2111.11638}}
  (\bibinfo{year}{2021}).
\newblock


\bibitem[Srivastava et~al\mbox{.}(2014)]%
        {srivastava2014dropout}
\bibfield{author}{\bibinfo{person}{Nitish Srivastava},
  \bibinfo{person}{Geoffrey Hinton}, \bibinfo{person}{Alex Krizhevsky},
  \bibinfo{person}{Ilya Sutskever}, {and} \bibinfo{person}{Ruslan
  Salakhutdinov}.} \bibinfo{year}{2014}\natexlab{}.
\newblock \showarticletitle{Dropout: a simple way to prevent neural networks
  from overfitting}.
\newblock \bibinfo{journal}{\emph{The journal of machine learning research}}
  (\bibinfo{year}{2014}), \bibinfo{pages}{1929--1958}.
\newblock


\bibitem[Veli{\v{c}}kovi{\'{c}} et~al\mbox{.}(2018)]%
        {velivckovic2017graph}
\bibfield{author}{\bibinfo{person}{Petar Veli{\v{c}}kovi{\'{c}}},
  \bibinfo{person}{Guillem Cucurull}, \bibinfo{person}{Arantxa Casanova},
  \bibinfo{person}{Adriana Romero}, \bibinfo{person}{Pietro Li{\`{o}}}, {and}
  \bibinfo{person}{Yoshua Bengio}.} \bibinfo{year}{2018}\natexlab{}.
\newblock \showarticletitle{Graph Attention Networks}. In
  \bibinfo{booktitle}{\emph{International Conference on Learning
  Representations}}.
\newblock


\bibitem[Wang et~al\mbox{.}(2023)]%
        {wang2023neural}
\bibfield{author}{\bibinfo{person}{Xiyuan Wang}, \bibinfo{person}{Haotong
  Yang}, {and} \bibinfo{person}{Muhan Zhang}.} \bibinfo{year}{2023}\natexlab{}.
\newblock \showarticletitle{Neural Common Neighbor with Completion for Link
  Prediction}.
\newblock \bibinfo{journal}{\emph{arXiv preprint arXiv:2302.00890}}
  (\bibinfo{year}{2023}).
\newblock


\bibitem[Wang and Zhang(2021)]%
        {wang2021glass}
\bibfield{author}{\bibinfo{person}{Xiyuan Wang} {and} \bibinfo{person}{Muhan
  Zhang}.} \bibinfo{year}{2021}\natexlab{}.
\newblock \showarticletitle{GLASS: GNN with Labeling Tricks for Subgraph
  Representation Learning}. In \bibinfo{booktitle}{\emph{International
  Conference on Learning Representations}}.
\newblock


\bibitem[Wang et~al\mbox{.}(2022)]%
        {wang2022towards}
\bibfield{author}{\bibinfo{person}{Zhaohui Wang}, \bibinfo{person}{Qi Cao},
  \bibinfo{person}{Huawei Shen}, \bibinfo{person}{Xu Bingbing},
  \bibinfo{person}{Muhan Zhang}, {and} \bibinfo{person}{Xueqi Cheng}.}
  \bibinfo{year}{2022}\natexlab{}.
\newblock \showarticletitle{Towards Efficient and Expressive GNNs for Graph
  Classification via Subgraph-Aware Weisfeiler-Lehman}. In
  \bibinfo{booktitle}{\emph{Proceedings of the First Learning on Graphs
  Conference}}. \bibinfo{pages}{17:1--17:18}.
\newblock


\bibitem[Wang et~al\mbox{.}(2014)]%
        {wang2014friendbook}
\bibfield{author}{\bibinfo{person}{Zhibo Wang}, \bibinfo{person}{Jilong Liao},
  \bibinfo{person}{Qing Cao}, \bibinfo{person}{Hairong Qi}, {and}
  \bibinfo{person}{Zhi Wang}.} \bibinfo{year}{2014}\natexlab{}.
\newblock \showarticletitle{Friendbook: a semantic-based friend recommendation
  system for social networks}.
\newblock \bibinfo{journal}{\emph{IEEE transactions on mobile computing}}
  (\bibinfo{year}{2014}), \bibinfo{pages}{538--551}.
\newblock


\bibitem[Xu et~al\mbox{.}(2019)]%
        {xu2018powerful}
\bibfield{author}{\bibinfo{person}{Keyulu Xu}, \bibinfo{person}{Weihua Hu},
  \bibinfo{person}{Jure Leskovec}, {and} \bibinfo{person}{Stefanie Jegelka}.}
  \bibinfo{year}{2019}\natexlab{}.
\newblock \showarticletitle{How Powerful are Graph Neural Networks?}. In
  \bibinfo{booktitle}{\emph{International Conference on Learning
  Representations}}.
\newblock


\bibitem[Xu et~al\mbox{.}(2018)]%
        {pmlr-v80-xu18c}
\bibfield{author}{\bibinfo{person}{Keyulu Xu}, \bibinfo{person}{Chengtao Li},
  \bibinfo{person}{Yonglong Tian}, \bibinfo{person}{Tomohiro Sonobe},
  \bibinfo{person}{Ken-ichi Kawarabayashi}, {and} \bibinfo{person}{Stefanie
  Jegelka}.} \bibinfo{year}{2018}\natexlab{}.
\newblock \showarticletitle{Representation Learning on Graphs with Jumping
  Knowledge Networks}. In \bibinfo{booktitle}{\emph{Proceedings of the 35th
  International Conference on Machine Learning}}. \bibinfo{pages}{5453--5462}.
\newblock


\bibitem[Yin et~al\mbox{.}(2022)]%
        {yin2022algorithm}
\bibfield{author}{\bibinfo{person}{Haoteng Yin}, \bibinfo{person}{Muhan Zhang},
  \bibinfo{person}{Yanbang Wang}, \bibinfo{person}{Jianguo Wang}, {and}
  \bibinfo{person}{Pan Li}.} \bibinfo{year}{2022}\natexlab{}.
\newblock \showarticletitle{Algorithm and System Co-design for Efficient
  Subgraph-based Graph Representation Learning}.
\newblock \bibinfo{journal}{\emph{arXiv preprint arXiv:2202.13538}}
  (\bibinfo{year}{2022}).
\newblock


\bibitem[Ying et~al\mbox{.}(2018)]%
        {ying2018graph}
\bibfield{author}{\bibinfo{person}{Rex Ying}, \bibinfo{person}{Ruining He},
  \bibinfo{person}{Kaifeng Chen}, \bibinfo{person}{Pong Eksombatchai},
  \bibinfo{person}{William~L. Hamilton}, {and} \bibinfo{person}{Jure
  Leskovec}.} \bibinfo{year}{2018}\natexlab{}.
\newblock \showarticletitle{Graph Convolutional Neural Networks for Web-Scale
  Recommender Systems}. In \bibinfo{booktitle}{\emph{Proceedings of the 24th
  ACM SIGKDD International Conference on Knowledge Discovery \& Data Mining}}.
  \bibinfo{pages}{974–983}.
\newblock
\showISBNx{9781450355520}


\bibitem[Zeng et~al\mbox{.}(2021)]%
        {zeng2021decoupling}
\bibfield{author}{\bibinfo{person}{Hanqing Zeng}, \bibinfo{person}{Muhan
  Zhang}, \bibinfo{person}{Yinglong Xia}, \bibinfo{person}{Ajitesh Srivastava},
  \bibinfo{person}{Andrey Malevich}, \bibinfo{person}{Rajgopal Kannan},
  \bibinfo{person}{Viktor Prasanna}, \bibinfo{person}{Long Jin}, {and}
  \bibinfo{person}{Ren Chen}.} \bibinfo{year}{2021}\natexlab{}.
\newblock \showarticletitle{Decoupling the depth and scope of graph neural
  networks}.
\newblock \bibinfo{journal}{\emph{Advances in Neural Information Processing
  Systems}} (\bibinfo{year}{2021}), \bibinfo{pages}{19665--19679}.
\newblock


\bibitem[Zhang and Chen(2018)]%
        {zhang2018link}
\bibfield{author}{\bibinfo{person}{Muhan Zhang} {and} \bibinfo{person}{Yixin
  Chen}.} \bibinfo{year}{2018}\natexlab{}.
\newblock \showarticletitle{Link Prediction Based on Graph Neural Networks}. In
  \bibinfo{booktitle}{\emph{Proceedings of the 32nd International Conference on
  Neural Information Processing Systems}}. \bibinfo{pages}{5171–5181}.
\newblock


\bibitem[Zhang et~al\mbox{.}(2018)]%
        {zhang2018end}
\bibfield{author}{\bibinfo{person}{Muhan Zhang}, \bibinfo{person}{Zhicheng
  Cui}, \bibinfo{person}{Marion Neumann}, {and} \bibinfo{person}{Yixin Chen}.}
  \bibinfo{year}{2018}\natexlab{}.
\newblock \showarticletitle{An End-to-End Deep Learning Architecture for Graph
  Classification}.
\newblock \bibinfo{journal}{\emph{Proceedings of the AAAI Conference on
  Artificial Intelligence}} (\bibinfo{year}{2018}).
\newblock


\bibitem[Zhang and Li(2021)]%
        {zhang2021nested}
\bibfield{author}{\bibinfo{person}{Muhan Zhang} {and} \bibinfo{person}{Pan
  Li}.} \bibinfo{year}{2021}\natexlab{}.
\newblock \showarticletitle{Nested Graph Neural Networks}. In
  \bibinfo{booktitle}{\emph{Advances in Neural Information Processing
  Systems}}. \bibinfo{pages}{15734--15747}.
\newblock


\bibitem[Zhang et~al\mbox{.}(2021)]%
        {zhang2021labeling}
\bibfield{author}{\bibinfo{person}{Muhan Zhang}, \bibinfo{person}{Pan Li},
  \bibinfo{person}{Yinglong Xia}, \bibinfo{person}{Kai Wang}, {and}
  \bibinfo{person}{Long Jin}.} \bibinfo{year}{2021}\natexlab{}.
\newblock \showarticletitle{Labeling Trick: A Theory of Using Graph Neural
  Networks for Multi-Node Representation Learning}. In
  \bibinfo{booktitle}{\emph{Advances in Neural Information Processing
  Systems}}. \bibinfo{pages}{9061--9073}.
\newblock


\end{thebibliography}

\end{document}